\documentclass[10pt,twocolumn,letterpaper]{article}
\usepackage[letterpaper,margin=0.72in,columnsep=0.28in]{geometry}

\usepackage{iftex}
\ifPDFTeX
  \usepackage[utf8]{inputenc}
  \usepackage[T1]{fontenc}
  \usepackage{tgpagella}
  \usepackage{textcomp}
  \usepackage{amssymb}          
\else
  \usepackage{fontspec}
  \usepackage{unicode-math}
  \IfFontExistsTF{texgyrepagella-math.otf}
    {\setmathfont{TeX Gyre Pagella Math}}
    {\setmathfont{Latin Modern Math}}
\fi
\usepackage{newunicodechar}

\usepackage{amsmath}
\usepackage{booktabs}
\usepackage{float}
\usepackage{graphicx}
\usepackage{adjustbox}
\usepackage{rotating}
\newlength{\rotmaxw}
\newlength{\rotmaxh}
\usepackage[font=small,labelfont=bf,skip=6pt]{caption}
\usepackage{xcolor}
\usepackage[verbose=silent]{microtype}
\usepackage{titlesec}
\usepackage{fancyhdr}
\usepackage{pifont}

\definecolor{accent}{RGB}{38,78,120}
\definecolor{linkblue}{RGB}{20,60,110}
\definecolor{urlblue}{RGB}{0,84,166}
\definecolor{headtint}{RGB}{24,42,64}

\usepackage[colorlinks=true,
            linkcolor=linkblue,
            citecolor=linkblue,
            urlcolor=urlblue,
            filecolor=urlblue,
            breaklinks=true]{hyperref}
\usepackage{xurl}

\ifPDFTeX
  \newcommand{\umap}[3]{\DeclareUnicodeCharacter{#2}{#3}}
\else
  \newcommand{\umap}[3]{\newunicodechar{#1}{#3}}
\fi
\umap{∈}{2208}{\ensuremath{\in}}
\umap{∉}{2209}{\ensuremath{\notin}}
\umap{∖}{2216}{\ensuremath{\setminus}}
\umap{∪}{222A}{\ensuremath{\cup}}
\umap{∩}{2229}{\ensuremath{\cap}}
\umap{⊆}{2286}{\ensuremath{\subseteq}}
\umap{⊂}{2282}{\ensuremath{\subset}}
\umap{ℝ}{211D}{\ensuremath{\mathbb{R}}}
\umap{∅}{2205}{\ensuremath{\emptyset}}
\umap{×}{00D7}{\ensuremath{\times}}
\umap{≤}{2264}{\ensuremath{\leq}}
\umap{≥}{2265}{\ensuremath{\geq}}
\umap{≈}{2248}{\ensuremath{\approx}}
\umap{≠}{2260}{\ensuremath{\neq}}
\umap{→}{2192}{\ensuremath{\rightarrow}}
\umap{↔}{2194}{\ensuremath{\leftrightarrow}}
\umap{−}{2212}{\ensuremath{-}}
\umap{±}{00B1}{\ensuremath{\pm}}
\umap{ρ}{03C1}{\ensuremath{\rho}}
\umap{τ}{03C4}{\ensuremath{\tau}}
\umap{Δ}{0394}{\ensuremath{\Delta}}
\umap{α}{03B1}{\ensuremath{\alpha}}
\umap{β}{03B2}{\ensuremath{\beta}}
\umap{²}{00B2}{\ensuremath{^{2}}}
\umap{³}{00B3}{\ensuremath{^{3}}}
\umap{₁}{2081}{\ensuremath{_{1}}}
\umap{₂}{2082}{\ensuremath{_{2}}}
\umap{·}{00B7}{\,\textperiodcentered\,}
\umap{✓}{2713}{\ding{51}}
\umap{✗}{2717}{\ding{55}}
\umap{ }{2009}{\,}   
\umap{ }{00A0}{~}   
\umap{‑}{2011}{-}   

\renewcommand{\arraystretch}{1.3}

\newcommand{\secsquare}{%
  \textcolor{accent}{\rule[0.14em]{0.52em}{0.52em}}\hspace{0.42em}}
\titleformat{\section}
  {\normalfont\large\bfseries\color{headtint}}{\secsquare\thesection}{0.45em}{}
\titlespacing*{\section}{0pt}{16pt plus 4pt minus 2pt}{5pt plus 1pt}
\titleformat{\subsection}
  {\normalfont\normalsize\bfseries\color{headtint}}{\thesubsection}{0.45em}{}
\titlespacing*{\subsection}{0pt}{11pt plus 3pt minus 2pt}{3pt}
\titleformat{\paragraph}[runin]
  {\normalfont\bfseries\color{headtint}}{}{0pt}{}[.]
\titlespacing*{\paragraph}{0pt}{6pt plus 2pt}{0.6em}

\newcommand{\shorttitle}{Auditing world-model explanations in retail forecasting}
\renewcommand{\headrulewidth}{0.4pt}
\renewcommand{\headrule}{\hbox to\headwidth{\color{accent!45}\leaders\hrule height \headrulewidth\hfill}}
\newcommand{\zEpisodes}{54}
\newcommand{\zScored}{38}
\newcommand{\zSlope}{-0.33}

\newcommand{\zSignShare}{63}

\newcommand{\zSizeLo}{6}
\newcommand{\zSizeHi}{10}

\newcommand{\zSlopeLo}{-0.68}
\newcommand{\zSlopeHi}{0.01}
\newcommand{\zFullSlope}{-0.52}
\newcommand{\zFullLo}{-0.81}
\newcommand{\zFullHi}{-0.23}
\newcommand{\zPlaceboSlope}{-0.26}
\newcommand{\zPlaceboLo}{-0.72}
\newcommand{\zPlaceboHi}{0.19}
\newcommand{\zPlaceboSig}{42}
\newcommand{\zPlaceboN}{38}
\newcommand{\zEarlyShare}{47}
\newcommand{\zEpisodeSE}{0.02}
\newcommand{\zPlaceboSD}{0.12}

\newcommand{\zCatN}{3}
\newcommand{\zStoresLo}{4}
\newcommand{\zStoresHi}{25}
\newcommand{\zControlsLo}{3}
\newcommand{\zControlsHi}{24}
\newcommand{\zSensLo}{26}
\newcommand{\zSensHi}{45}
\newcommand{\zSensSlopeLo}{-0.38}
\newcommand{\zSensSlopeHi}{-0.16}

\newcommand{\rVersions}{494}
\newcommand{\rWindows}{933}
\newcommand{\rVerM}{31}

\newcommand{\rPassMCb}{22 of 31}

\newcommand{\rMedMCeRank}{0.11}
\newcommand{\rMedMCeBias}{1.00}
\newcommand{\rMedMCf}{0.40}
\newcommand{\rMedMCdPct}{49}

\newcommand{\rVerD}{279}

\newcommand{\rPassDCa}{68 of 540}
\newcommand{\rPassDCb}{423 of 540}

\newcommand{\rMedDCeRank}{0.64}
\newcommand{\rMedDCeBias}{0.98}
\newcommand{\rMedDCf}{0.39}
\newcommand{\rMedDCdPct}{49}

\newcommand{\rVerF}{184}
\newcommand{\rWinF}{362}
\newcommand{\rPassFCa}{38 of 362}
\newcommand{\rPassFCb}{325 of 362}

\newcommand{\rAllF}{35}

\newcommand{\rMedFCeRank}{0.52}
\newcommand{\rMedFCeBias}{1.03}
\newcommand{\rMedFCf}{0.73}
\newcommand{\rMedFCdPct}{0}

\newcommand{\rCaPartsMax}{2e-16}
\newcommand{\rCaBaseMax}{1.7}
\newcommand{\rStabLiftM}{0.66}
\newcommand{\rStabFcM}{0.98}

\newcommand{\rStabLiftD}{0.85}

\newcommand{\rStabLiftF}{0.96}
\newcommand{\rStabFcF}{0.99}

\newcommand{\rRhoPassedM}{$-$0.32}
\newcommand{\rRhoPassedD}{$-$0.19}
\newcommand{\rRhoPassedF}{0.26}
\newcommand{\rRhoPassedPool}{$-$0.08}
\newcommand{\rPPassed}{0.205}
\newcommand{\rRhoPassedPoolV}{$-$0.19}
\newcommand{\rPPassedV}{0.004}

\newcommand{\rRhoCeRankPool}{$-$0.42}

\newcommand{\rRhoCeBiasPool}{0.45}

\newcommand{\rWwRho}{$-$0.10}
\newcommand{\rWwLo}{$-$0.21}
\newcommand{\rWwHi}{0.02}
\newcommand{\rWwP}{0.10}

\newcommand{\rWwNoCeRho}{$-$0.02}
\newcommand{\rWwNoCeLo}{$-$0.13}
\newcommand{\rWwNoCeHi}{0.12}

\newcommand{\rWwNoCaCeRho}{$-$0.12}
\newcommand{\rWwNoCaCeLo}{$-$0.24}
\newcommand{\rWwNoCaCeHi}{0.03}

\newcommand{\rWwConfigs}{147}
\newcommand{\rWwData}{16}
\newcommand{\rZEpisodes}{38}
\newcommand{\rZWorldInside}{22}
\newcommand{\rZLgbInside}{13}
\newcommand{\rZWorldSign}{32}
\newcommand{\rZLgbSign}{31}
\newcommand{\rZMeasuredSign}{25}
\newcommand{\rZSlopeMeas}{$-$0.33}
\newcommand{\rZSlopeMeasLo}{$-$0.68}
\newcommand{\rZSlopeMeasHi}{0.01}
\newcommand{\rZSlopeWorld}{$-$0.62}
\newcommand{\rZSlopeWorldLo}{$-$0.80}
\newcommand{\rZSlopeWorldHi}{$-$0.44}
\newcommand{\rZSlopeLgb}{$-$0.75}
\newcommand{\rZSlopeLgbLo}{$-$0.92}
\newcommand{\rZSlopeLgbHi}{$-$0.58}

\newcommand{\tSets}{nine}

\newcommand{\tCThreeLo}{0.71}
\newcommand{\tCThreeHi}{0.98}
\newcommand{\tCSixLo}{0.33}
\newcommand{\tCSixHi}{0.75}
\newcommand{\tReadGainLo}{0.17}
\newcommand{\tReadGainHi}{0.91}
\newcommand{\tReadRolledLo}{0.85}
\newcommand{\tReadRolledHi}{0.99}
\newcommand{\tSlopeMeas}{-0.33}
\newcommand{\tSlopeMeasLo}{-0.68}
\newcommand{\tSlopeMeasHi}{0.01}
\newcommand{\tSlopeWm}{-0.73}
\newcommand{\tSlopeWmLo}{-0.93}
\newcommand{\tSlopeWmHi}{-0.54}
\newcommand{\tSlopeLgb}{-0.63}
\newcommand{\tSlopeLgbLo}{-0.79}
\newcommand{\tSlopeLgbHi}{-0.47}
\newcommand{\tRhoAll}{0.02}
\newcommand{\tPAll}{0.444}
\newcommand{\tRhoAllLo}{-0.02}
\newcommand{\tRhoAllHi}{0.05}
\newcommand{\tConfigsAll}{703}

\newcommand{\tRhoNoOneFive}{0.03}
\newcommand{\tPNoOneFive}{0.101}

\newcommand{\tSearchVersions}{2912}
\newcommand{\tCFourDomLo}{29}
\newcommand{\tCFourDomHi}{83}
\newcommand{\tCFourMfive}{18}
\newcommand{\tCFourFav}{49}
\newcommand{\tCFourFrat}{0.02}
\newcommand{\tCFiveRankMfive}{0.18}
\newcommand{\tWindows}{48}
\newcommand{\tCOneWin}{5 of 48}
\newcommand{\tCOneWinPct}{10}

\newcommand{\tCTwoWin}{39 of 48}
\newcommand{\tCTwoWinPct}{81}
\newcommand{\tCThreeWin}{41 of 44}
\newcommand{\tCThreeWinPct}{93}

\newcommand{\tCFiveWin}{36 of 48}
\newcommand{\tCFiveWinPct}{75}
\newcommand{\tCSixWin}{18 of 48}
\newcommand{\tCSixWinPct}{38}

\newcommand{\dRuns}{14}
\newcommand{\dWindows}{24}

\newcommand{\dCdAllD}{73}
\newcommand{\dCdTenthD}{73}
\newcommand{\dCdBigD}{57}

\newcommand{\dCdAllM}{46}
\newcommand{\dCdTenthM}{45}
\newcommand{\dCdBigM}{11}

\newcommand{\cIfOf}{933}
\newcommand{\cIfSameN}{37}

\newcommand{\cIfDiffN}{896}

\newcommand{\cIfFailN}{796}

\newcommand{\cIfAnyM}{0 of 31}

\newcommand{\cIfAnyD}{534 of 540}

\newcommand{\cIfAnyF}{362 of 362}

\newcommand{\cIfShareNoReg}{37}
\newcommand{\cIfShareRule}{61}
\newcommand{\cIfShareNew}{2}

\newcommand{\cIfShareOne}{34}

\IfFileExists{generated/known_macros.tex}{\newcommand{\kaClSeedRank}{0.99}

\newcommand{\kaClWmTruth}{0.99}
\newcommand{\kaClWmBias}{1.14}

\newcommand{\kaClWmRmsse}{0.401}

\newcommand{\kaClWmNpCfour}{31.5}

\newcommand{\kaClWmNlTruth}{0.89}

\newcommand{\kaClWmNlPriceBias}{1.35}
\newcommand{\kaClWmNlPromoBias}{0.14}
\newcommand{\kaClWmNlDisplayBias}{0.05}
\newcommand{\kaClLgbTruth}{0.85}
\newcommand{\kaClLgbBias}{1.09}

\newcommand{\kaClLgbCfour}{7.6}

\newcommand{\kaClLgbNpCfour}{49.5}

\newcommand{\kaCoWmRmsse}{0.345}
\newcommand{\kaCoWmCfive}{1.06}

\newcommand{\kaCoWmNpCfour}{27.2}

\newcommand{\kaCoLgbBias}{1.26}

\newcommand{\kaCoLgbCfour}{9.9}

\newcommand{\kaCoLgbNpCfour}{43.4}

\newcommand{\kaPassTruthRho}{0.87}
\newcommand{\kaPassBiasRho}{−0.27}

\newcommand{\kaEncLo}{0.44}
\newcommand{\kaEncHi}{0.70}
\newcommand{\kaLiftReadLo}{0.15}
\newcommand{\kaLiftReadHi}{0.82}

\newcommand{\kaClMeasSlope}{−0.73}
\newcommand{\kaClMeasLo}{−1.41}
\newcommand{\kaClMeasHi}{−0.06}

\newcommand{\kaClWmCsixPromo}{0.80}

\newcommand{\kaClWmCsevenLo}{−2.42}
\newcommand{\kaClWmCsevenHi}{−2.37}
\newcommand{\kaClWmTimesLo}{4.7}

\newcommand{\kaClNlCsixPromo}{0.29}
\newcommand{\kaClNlCsixDisplay}{−0.17}
\newcommand{\kaClNlCsix}{0.67}
\newcommand{\kaClLgbCsevenSlope}{−0.64}

\newcommand{\kaCoMeasSlope}{−0.74}
\newcommand{\kaCoMeasLo}{−1.90}
\newcommand{\kaCoMeasHi}{0.41}
\newcommand{\kaCoWmBiasLo}{1.33}
\newcommand{\kaCoWmBiasHi}{1.40}
\newcommand{\kaCoWmDispLo}{1.81}
\newcommand{\kaCoWmDispHi}{1.93}

\newcommand{\kaCoWmCsix}{0.75}

\newcommand{\kaCoWmCsevenLo}{−3.25}
\newcommand{\kaCoWmCsevenHi}{−3.12}

\newcommand{\kaCoWmTimesHi}{6.5}

\newcommand{\kaCoNlCsix}{0.54}
\newcommand{\kaCoLgbCsevenSlope}{−0.70}

\newcommand{\kaPlanned}{1,743}
\newcommand{\kaRows}{2,880}
}{}
\newcommand{\ksN}{eight}
\newcommand{\ksWmBiasLo}{1.23}
\newcommand{\ksWmBiasHi}{1.42}

\newcommand{\ksWmRegLo}{3.0}
\newcommand{\ksWmRegHi}{12.5}
\newcommand{\ksLgbBiasLo}{1.16}
\newcommand{\ksLgbBiasHi}{1.35}
\newcommand{\ksLgbRegLo}{0.7}
\newcommand{\ksLgbRegHi}{3.0}

\newcommand{\ksWmPassedHi}{4}
\newcommand{\ksWmApplicable}{5}
\newcommand{\ksWmStockLo}{0.34}
\newcommand{\ksWmStockHi}{0.52}
\newcommand{\kdClWmChoiceLo}{0.996}
\newcommand{\kdClWmChoiceHi}{0.998}
\newcommand{\kdClWmStockLo}{0.34}
\newcommand{\kdClWmStockHi}{0.35}
\newcommand{\kdClLgbChoice}{0.965}
\newcommand{\kdClLgbStock}{0.56}
\newcommand{\kdClRandom}{0.41}

\newcommand{\kdCoWmChoiceLo}{0.994}
\newcommand{\kdCoWmChoiceHi}{0.997}
\newcommand{\kdCoWmStockLo}{0.44}
\newcommand{\kdCoWmStockHi}{0.48}
\newcommand{\kdCoLgbChoice}{0.971}
\newcommand{\kdCoLgbStock}{0.50}
\newcommand{\kdCoRandom}{0.45}

\newcommand{\cfoAllDomLo}{29}
\newcommand{\cfoAllDomHi}{83}
\newcommand{\cfoObsDomLo}{23}
\newcommand{\cfoObsDomHi}{80}
\newcommand{\cfoKeptDomLo}{57}
\newcommand{\cfoKeptDomHi}{79}

\newcommand{\cfoAllMfive}{18}
\newcommand{\cfoKeptMfive}{26}
\newcommand{\cfoObsMfive}{19}

\ifdefined\tmlrtrue\else\expandafter\newif\csname iftmlr\endcsname\fi
\newcommand{\companion}{\iftmlr forecasting study\else companion paper\fi}
\newcommand{\ourcompanion}{\iftmlr the forecasting study\else our companion paper\fi}

\begin{document}

\twocolumn[
  \vspace*{6pt}
  \begin{center}
    {\LARGE\bfseries\color{headtint} Auditing World-Model Explanations in Retail Forecasting:\\[2pt]
     Faithfulness, Accuracy and Price Response\par}
    \vspace{10pt}
    {\large Karthik Iyer\par}
    \vspace{3pt}
    {\footnotesize Correspondence:
     \href{mailto:karthik.iyer.au@gmail.com}{karthik.iyer.au@gmail.com}
     \quad ORCID:
     \href{https://orcid.org/0009-0004-0593-1602}{0009-0004-0593-1602}\par}
    \vspace{18pt}
    \begin{minipage}{0.86\textwidth}
      \iftmlr\begin{abstract}\else\small
\textbf{Abstract.}
\fi
Forecasting models that read the price and promotion plan explain each forecast
as a baseline, what would sell with nothing planned, plus the lift the plan
adds, and merchants order stock and plan promotions on that lift. We show that
an explanation can pass the checks used to trust it and still be wrong enough
to cost money. We define eight checks for any forecaster that states a plan
lift, six with a pass rule fixed before any result, and run them on a world
model and on LightGBM. In generated store data with known effects, the world
model passes the checks of construction, stability, price direction and
agreement with sales and with LightGBM, yet when merchants plan into
high-demand weeks it states \kaCoWmBiasLo{} to \kaCoWmBiasHi{} times the true
lift, and \ksWmBiasLo{} to \ksWmBiasHi{} times in \ksN{} variants of that
world; LightGBM overstates it too (\kaCoLgbBias{}, and \ksLgbBiasLo{} to
\ksLgbBiasHi{}). Both still pick the right weeks to promote, but stock ordered
to the world model's lift is off by \kdCoWmStockLo{} to \kdCoWmStockHi{} of the
true lift. The world model's stated response to a lasting regular-price change
is \kaClWmTimesLo{} to \kaCoWmTimesHi{} times the true $-0.5$, an error in kind
that only a measured cause finds. Sanity checks for explanations test whether an
explanation depends on the model; a decision needs to know whether it is right
about the world, and these results show the first does not give the second. For
the final tuned world models of \ourcompanion{} on \tSets{} data sets, baseline
and lift add up under each model's own plan on every window, but a deeper price
cut lowers the forecast in \tCFourDomLo\% to \tCFourDomHi\% of item-week steps on
Dominick's. Across \tConfigsAll{} configurations from the settings searches,
checks passed were not related to accuracy (Spearman $\tRhoAll$, 95\% interval
$\tRhoAllLo$ to $\tRhoAllHi$). A noisy benchmark of price episodes in
Dominick's data is reported as a check of direction only.

\iftmlr\end{abstract}\else\vspace{4pt}\fi

    \end{minipage}
    \vspace{16pt}
  \end{center}
]
\thispagestyle{plain}

\section{Introduction}
\label{sec:intro}

A retail forecast is used for more than its number. When a merchant plans a
promotion, the question is not only how much will sell but how much of it the
promotion will add, and whether a deeper price cut would add more. Forecasting
models that see the plan can answer that directly: they forecast the item with
the plan as made, and again with nothing planned, and the difference is the lift
the plan adds. The world model and the lift model of \ourcompanion{}
~\cite{rwm} are built this way: each forecast is a baseline under the
nothing-planned plan plus a lift, and the lift is exactly zero when nothing is
planned. The split is what a merchant reads ("the promotion added 40 units"),
plans on, and is judged by.

Such an explanation is useful only if it can be trusted, and being built into
the model does not make it right. The explanation literature separates two
properties~\cite{jacovi2020,chen2020}: an explanation is \emph{faithful} if it
reflects what the model does, and \emph{correct} if it reflects what happens in
the world. In molecular property prediction, MolSanity found the two come
apart: attributions faithful to the model were often anti-aligned with the
substructure that determines the property~\cite{molsanity}. Retail has a
property chemistry lacks. Some causes can be measured: Dominick's Finer Foods
moved store prices on purpose in the early 1990s~\cite{hoch1994}, so the
response of sales to a price change can be estimated from the data and a
model's stated response compared with it.

This paper audits plan-lift explanations in retail forecasting. It asks three
questions, fixed in a plan before any result~\cite{plan2}: are the
explanations faithful to the model, stable and sensible; do they match what
sold and what a price change causes; and across model versions, do versions
whose explanations pass more checks also forecast better?

\begin{figure*}[t]
\centering
\includegraphics[width=\textwidth]{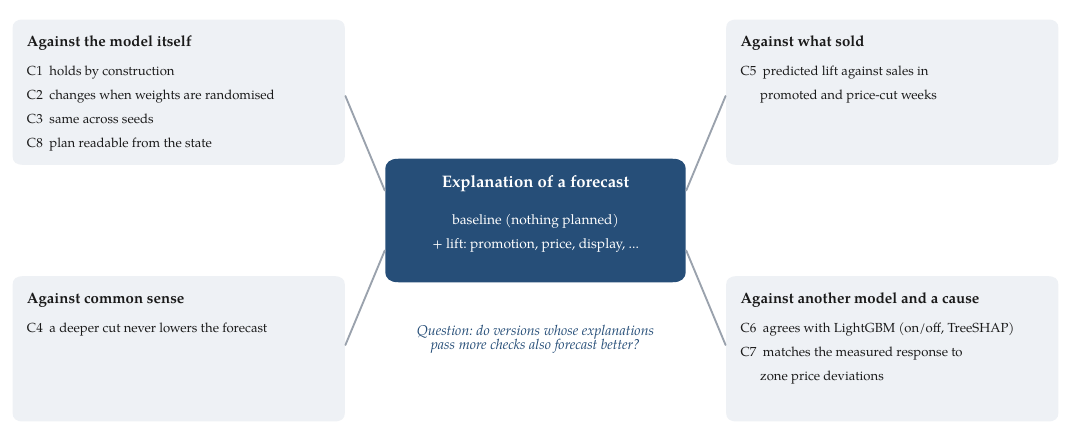}
\caption{\textbf{Each forecast's explanation is checked against the model
itself, common sense, what sold, another model and a measured cause.} The
explanation is reported in units, baseline plus lift. Pass rules for C1 to C6
are fixed in \texttt{docs/PAPER2\_PLAN.md} before any result; C7 and C8 are
reported per model (Table~\ref{tab:thresholds}). Every model version and every
check is reported.}
\label{fig:audit}
\end{figure*}

\paragraph{Contributions}
\begin{enumerate}\setlength{\itemsep}{1pt}
\item A definition of the plan-lift explanation of a forecast that applies to
any model that reads the plan, whatever its architecture, and eight checks on
it, run here on a world model and on LightGBM
(Figure~\ref{fig:audit}), six of them (C1 to C6) with a pass rule fixed before
any result.
\item A known-answer study: generated store data in which the effect of every
part of the plan is known, with failures planted on purpose, to measure which
checks catch which failure before they are trusted on real data.
\item A noisy observational benchmark on public retail data: \zEpisodes{}
episodes in six Dominick's categories in which one store's regular prices sat
apart from its zone peers, \zScored{} of them scored, with their effect on
sales measured by difference in differences, an event study, a placebo and a
sensitivity analysis. Which stores were treated is inferred, prices move before
the detected start, and a placebo at false dates gives a slope close to the
estimate, so it is not a ground truth.
\item An audit of the final tuned world model the \companion{}  reports for
each of its \tSets{} data sets, against the tuned LightGBM, and a test of
whether explanations that pass more checks go with more accurate forecasts. An
earlier audit of the \companion{} 's development versions, made in October
2026, is kept as a record (\S\ref{sec:results}).
\end{enumerate}

\paragraph{Findings}
The finding that matters for practice is that the checks commonly used to
trust an explanation, that it depends on the model, adds up and points the
right way, do not show that it is right about the world, and a merchant acts on
the world. In generated data with known effects, explanations that pass the checks against
the model, the data and another model can still be wrong in size, when
merchants plan into strong weeks (for the world model and for LightGBM alike), and wrong in kind, when a model answers a
question about the regular price with a response close to what it learned from
temporary cuts. The price-direction check and a lever-by-lever comparison with
a second model catch the cruder failures; the randomisation check does not test
an explanation built into the model; and only a measured cause finds the error
in kind (\S\ref{sec:known}). On Dominick's, zone price deviations give a pooled
8-week response close to a placebo at false dates and a larger full-episode
response whose interval excludes zero; single episodes carry no usable
information (\S\ref{sec:zones}). The final tuned world models
(\S\ref{sec:tuned}) pass C1 under their own plan for nothing planned on every
window, but a deeper cut lowers the forecast in \tCFourDomLo\% to
\tCFourDomHi\% of item-week steps on Dominick's, and both they and LightGBM
state about twice the measured response to the zone price deviations. Across
the configurations the settings searches tried, the number of checks passed
was not related to accuracy; with no known answer for those versions, this is a finding about accuracy, and
whether an explanation is right rests on the known-answer study. These are findings for these models on these
public data sets.

\section{Related work}
\label{sec:related}

\paragraph{What an explanation should satisfy}
Interpretability lacks a single definition~\cite{lipton2018,doshivelez2017}.
Jacovi and Goldberg~\cite{jacovi2020} separate faithfulness, whether an
explanation reflects the model's reasoning, from plausibility, whether it
convinces a person, and warn that evaluations often measure the second. Chen et
al.~\cite{chen2020} draw the related line between explanations true to the
model and true to the data, and Janzing et al.~\cite{janzing2020} argue that
feature relevance is a causal question about interventions on the model's
inputs. Our on/off contributions are interventions of exactly that kind, made
on the plan inputs the merchant controls.

\paragraph{Feature attributions and their failures}
LIME~\cite{ribeiro2016} and SHAP~\cite{lundberg2017}, with its exact form for
trees~\cite{lundberg2020}, are the most used attributions~\cite{adadi2018}.
Saliency maps can barely change when a network's weights are
randomised~\cite{adebayo2018}, move with input shifts the model
ignores~\cite{kindermans2019}, differ for near-identical
inputs~\cite{alvarezmelis2018}, and rank features whose removal does not hurt
the model~\cite{hooker2019}. Rudin~\cite{rudin2019} argues for models that are
interpretable by construction. The world model and the lift model are of that
kind for the plan, which lets this audit compare a built-in explanation with
post-hoc ones on the same forecasts.

\paragraph{Explanations of time series and forecasts}
Rojat et al.~\cite{rojat2021} survey explanation methods for time series.
Ismail et al.~\cite{ismail2020} and Tonekaboni et
al.~\cite{tonekaboni2020} show on generated data with known important time steps
that saliency methods designed for images often fail on temporal data, the
approach our known-answer study takes for retail plans. Among forecasting
models, the Temporal Fusion Transformer~\cite{tft} reports attention weights
as explanations. We know of no study that checks the explanation of a demand
forecast against a measured promotional or price effect.

\paragraph{Promotion and price effects}
Marketing science has measured promotional effects for decades: the split of a
promotion's sales into brand switching, timing and quantity~\cite{gupta1988,
vanheerde2003}, the dip after a promotion~\cite{mace2004,vanheerde2000}, and
the size of price elasticities and what drives it~\cite{bijmolt2005}. Promotion forecasting systems built on these
effects include PromoCast~\cite{promocast}; Ma, Fildes and Huang~\cite{ma2016}
and Fildes et al.~\cite{fildes2022} review the forecasting side. These elasticities describe temporary cuts. Regular price changes move far fewer
units: in Dominick's randomised store-level pricing experiments~\cite{hoch1994},
regular prices were raised or lowered 10\% across many categories, and volume
moved about 3\% over the following 16 weeks, an elasticity near $-0.3$. Hoch et al.~\cite{hoch1995} and
Montgomery~\cite{montgomery1997} model store-level price response on the same
data.

\paragraph{Quasi-experiments in scanner data}
The public Dominick's files do not mark the experimental stores or weeks.
Abrams, Gui and Horta\c{c}su~\cite{abrams} reconstructed them with the
original documentation and found that some published price changes did not
replicate. We therefore treat the price deviations visible in the public data
as a quasi-experiment, never as the experiments, and estimate their effect by
difference in differences~\cite{angrist2009} with standard errors clustered by
store~\cite{bertrand2004,cameron2015}.

\section{The explanation and the checks}
\label{sec:checks}

\subsection{Notation}

A data set has series $i$ (an item in a store) observed in periods $t$. The
plan for series $i$ in period $t$ is $x_{it} = (p_{it}, \ell^{1}_{it}, \dots,
\ell^{L}_{it})$: the shelf price and $L$ lever indicators (a deal flag on
Dominick's; promotion, display, feature and a tag-only price reduction
flag on Frat; event and SNAP days on M5; promotion, event and payday days on
Favorita, which has no prices).
A model fitted on the periods before an origin forecasts the next $H$ periods
from the history and the plan for those periods, $\hat y_{it} = f_{it}(x)$.

The audit's \emph{regular price} $r_i$ is the series' highest price over the
last 12 training periods, or the planned price where the series has no price in
those periods. The plan with \emph{nothing planned}, $x^{0}$, sets every price
to $r_i$ and every lever to zero for all $H$ periods. Each model also has its
own regular price and its own plan for nothing planned, which can differ from
the audit's (\S\ref{sec:results}); that difference is what makes C1 fail under
the audit's plan. The explanation of a forecast has four parts:
\begin{align}
  b_{it} &= f_{it}(x^{0}) \qquad \text{baseline}\\
  \Lambda_{it} &= f_{it}(x) - b_{it} \qquad \text{total lift}\\
  c^{k}_{it} &= f_{it}(x) - f_{it}(x \mid \ell^{k} = 0) \qquad \text{lever } k\\
  c^{p}_{it} &= f_{it}(x) - f_{it}(x \mid p = r) \qquad \text{price}
\end{align}
and the interaction $\iota_{it} = \Lambda_{it} - c^{p}_{it} - \sum_k c^{k}_{it}$,
which is zero only when the levers act separately. All four are in units.
Switching a lever off switches it off for all $H$ periods, so the dip after a
promotion is part of that promotion's contribution.

The world model and the lift model of the \companion{} ~\cite{rwm} compute
the split themselves. Their network keeps a \emph{state} for each series: an
\emph{encoder} turns the history into the starting state, and a
\emph{transition} steps the state forward one period at a time given that
period's plan (the state is \emph{rolled} through the plan). Two output layers
read each period's state: the \emph{baseline readout} gives the forecast under
the model's own plan for nothing planned, and the \emph{lift readout} gives a
lift factor, the readout on the plan as made divided by the same readout on the
nothing-planned plan, which is exactly one when nothing is planned. The forecast
is the baseline times the lift factor; the explanation reports it in units, as
the baseline $b$ plus the lift $\Lambda = \hat y - b$. For
LightGBM~\cite{lightgbm} the same on/off quantities are computed from its
forecasts, together with TreeSHAP values $\phi^{k}_{it}$~\cite{lundberg2020}
on its lever and price inputs. Unlike Shapley values, on/off contributions
answer the merchant's question (what would sell without this lever) directly,
and are defined for any model.

Rank agreement $\rho(a, b)$ below is the Spearman correlation over item-weeks
where either $a$ or $b$ is non-zero, so that the many weeks with nothing planned
do not inflate it.

\subsection{The eight checks}

\paragraph{C1 By construction} Three violations, each relative to the largest
forecast: the lift with nothing planned, $\max|\Lambda(x^{0})|$; the movement
of the baseline with the plan, $\max|b(x) - b(x^{0})|$; and
$\max|b + \Lambda - \hat y|$, with $b$ and $\Lambda$ as the model's readouts
give them. These hold by design and are checked because a design can be broken
by an implementation.

\paragraph{C2 Randomisation} Following Adebayo et al.~\cite{adebayo2018}, the
network's parts are re-initialised in turn from the output back to the input:
lift readout, baseline readout, transition, encoder. After each step, the rank
agreement of the total lift with the original, $\rho(\Lambda, \Lambda^{(k)})$,
should fall. An explanation that survives randomisation does not depend on what
the model learned. The known-answer study shows that C2 does not test a lift
built into the model (\S\ref{sec:known}); from there on it is reported, and
counted in the main test as fixed in advance, but not used as evidence.

\paragraph{C3 Seed stability} Fits of the same configuration that differ only
in their seed should explain a forecast alike. The score is the mean over seed
pairs of $\rho(\Lambda^{s}, \Lambda^{s'})$, shown against the same agreement for
the forecasts themselves. For a final tuned model, which averages copies fitted
with different seeds, the pairs are its copies.

\paragraph{C4 Price direction} The planned price is cut by 5\%, 10\%, 20\% and
30\% in turn. For each item-week and each step to a deeper cut (0 to 5\%, 5\%
to 10\%, 10\% to 20\%, 20\% to 30\%), the step fails when the deeper cut gives
a forecast lower than the shallower one by more than 0.1\% of the forecast. The
score is the share of item-week steps that fail.

\paragraph{C5 Against what sold} On item-weeks with a lever on or a price at
least 5\% below regular, the bias $\sum \hat y / \sum y$ and the rank agreement
of the predicted lift factor $\hat y/b$ with the observed one $y/b$. Merchants
plan promotions for weeks they expect to sell well, so $y/b$ overstates what
the plan caused; this check measures agreement with the data, not with a cause.

\paragraph{C6 Against LightGBM} The rank agreement of the total lift with
LightGBM's on the same window, $\rho(\Lambda, \Lambda_{\mathrm{LGB}})$, and for
each lever $\rho(c^{k}, c^{k}_{\mathrm{LGB}})$ and $\rho(c^{k}, \phi^{k})$. Two
models fitted to the same data can agree and both be wrong; agreement is
evidence of a shared signal, not of correctness.

\paragraph{C7 Against a measured cause} For an episode $e$ in which store $s$'s
regular prices sat at a log ratio $\delta_e$ from its zone peers, the model's
stated response is
\begin{equation}
  \hat R_e = \log \frac{\sum_{t \in E} \sum_{i \in s} f_{it}(x)}
                      {\sum_{t \in E} \sum_{i \in s}
                       f_{it}(x \mid p_{s} = p_{s} e^{-\delta_e})},
\end{equation}
the change in the store's forecast category units if its prices had stayed with
its zone, over the episode's first eight weeks $E$. It is compared with the
measured response $\tau_e$ from difference in differences
(\S\ref{sec:zones}). The score fixed in advance is per episode: whether
$\hat R_e$ falls inside $\tau_e$'s 95\% interval and whether its sign agrees.
The comparison of the pooled slopes of $\hat R_e$ and $\tau_e$ on $\delta_e$
was added after the robustness results showed the per-episode intervals to be
too narrow, and is exploratory (Appendix~\ref{app:plan}).

\paragraph{C8 Plan read-back} For the world model only: a ridge probe reads
each plan input of a period (the price as a share of the regular and of the
usual price, each lever, and whether each is recorded) back from the model's
state after it has been rolled through the plan. The score is the $R^2$,
averaged over the inputs, against the same probe on the state before the
window, which has not seen the plan. Because the rolled state has received the
plan as input, a high $R^2$ shows only that the state keeps it; the gain over
the state before the window is the informative part.

\subsection{Pass thresholds and the main test}
\label{sec:test}

Pass rules for C1 to C6 were fixed in the plan before any audit result was read
(Table~\ref{tab:thresholds}); C7 and C8 have none and are reported per model.
A \emph{model version} is a configuration and a seed; a version-window is a
version scored on one rolling window. For the main question, each
version-window gets the number of checks it passes among C1 to C6, counting
only those that apply to it. Within each data set the Spearman correlation
across version-windows between that count and item-by-store WRMSSE (lower is
better) is computed, and the same for each check's continuous score. A negative
correlation means versions that pass more checks forecast better. The data sets
are pooled by a permutation test that shuffles the scores within each data set,
10{,}000 times.

\begin{table}[t]
\centering
\caption{\textbf{C1 to C6 have a pass rule fixed before any result; C7 and C8
do not.} Scores as defined in \S\ref{sec:checks}; a check that does not apply
to a model (C1 and C2 for LightGBM) is not counted for it.}
\label{tab:thresholds}
\footnotesize
\begin{tabular}{@{}lp{3.3cm}p{2.3cm}@{}}
\toprule
Check & Passes when & Applies to\\
\midrule
C1 & every violation $< 0.001$ & lift readout models\\
C2 & $\rho$ after the lift readout $< 0.5$ & neural models\\
C3 & mean seed $\rho \ge 0.7$ & configurations with $\ge 2$ seeds\\
C4 & share $< 1\%$ & all\\
C5 & bias in $[0.8, 1.25]$ and rank $\ge 0.3$ & all\\
C6 & $\rho(\Lambda, \Lambda_{\mathrm{LGB}}) \ge 0.5$ & non-LightGBM\\
C7 & no pass rule; per-episode score fixed in advance, pooled slope added after
the robustness results & Dominick's models\\
C8 & no pass rule & world model\\
\bottomrule
\end{tabular}
\end{table}

\section{A known-answer study}
\label{sec:known}

Before the checks are trusted on real data, they are run where the answer is
known. Following Ismail et al.~\cite{ismail2020} and Tonekaboni et
al.~\cite{tonekaboni2020}, who test time-series explanations on generated data
with known important inputs, we generate weekly store data in which the effect
of every part of the plan is set by us, fit correct models and models with
failures planted on purpose, and ask two questions: which checks catch which
failure, and do the explanations of models that pass the checks match the
truth?

\subsection{Design}

\begin{figure}[t]
\centering
\includegraphics[width=\columnwidth]{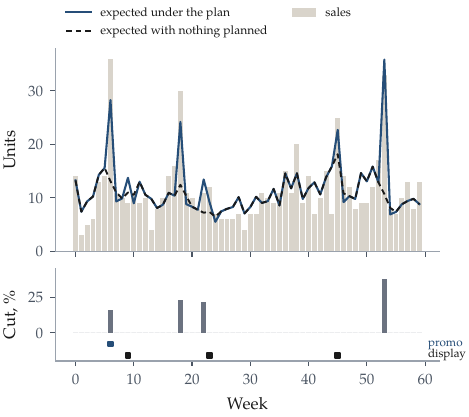}
\caption{\textbf{In the generated data the effect of every part of the plan is
known.} One item in one store over 60 weeks, in the world where merchants plan
into high-demand weeks: sales, expected sales under the plan, and expected
sales with nothing planned; below, the cut and the weeks with a promotion or a
display.}
\label{fig:knownseries}
\end{figure}

\paragraph{The data}
Three price zones of six stores sell 20 items for 260 weeks
(Figure~\ref{fig:knownseries}). Expected sales are the item's size and season
times an unobserved demand shock (a first-order autoregression on the log
scale, standard deviation 0.25, persistence 0.7), times the effects in
Table~\ref{tab:knowntruth}. A temporary cut and a change in the regular price
have different elasticities, $-2.5$ and $-0.5$, as in the literature on
promotions and the Dominick's experiments. Each item gets one chain-wide
regular price rise of 3\% to 8\%. In each zone one store's regular prices rise
10\% and one store's fall 10\% for 16 weeks at random times during training,
and one rises and one falls starting at the forecast origin, week 253, so that
every model can be scored on them exactly as on Dominick's. Sales are Poisson
around their expectation.

\paragraph{Two worlds}
In the first, cuts, promotions and displays are planned at random. In the
second, merchants plan them more often in weeks when the unobserved demand
shock is high (the log odds of each rise by the shock in standard deviations),
as merchants do when they promote into strong weeks. Everything else is the
same.

\paragraph{The models}
In each world: the world model with three seeds; the world model fitted with
each series' prices shuffled across its training weeks, so that it cannot learn
a price effect (\emph{blind to price}); the world model fitted with promotion
and display shuffled the same way (\emph{blind to levers}); LightGBM; and
LightGBM blind to price. The forecast for weeks 253 to 260 is explained and
every check run as on real data. The world model has the \companion{} 's
Dominick's world-model settings as they stood at the earlier audit, reduced to fit
on a processor: one fit rather than an average of copies, width 32, 2,000
training steps, and without the option that rescales item forecasts within a
store-week for shoppers switching between items. LightGBM has the \companion{}
's Dominick's settings. Each explanation is also scored against the
truth, computed from the expected sales under the same counterfactual plans:
rank agreement and the ratio of summed lifts, overall and by lever, on the
\kaPlanned{} of \kaRows{} item-weeks whose true lift is not zero. Besides the
weeks with a cut, a promotion or a display, these include the week after a
planned week, with its dip, and every week priced away from the regular price
the audit uses, such as the weeks of a regular price change.

\begin{table}[t]
\centering
\caption{\textbf{The generated data have known effects for every part of the
plan.}}
\label{tab:knowntruth}
\footnotesize
\begin{tabular}{@{}lr@{}}
\toprule
Effect on expected sales & Size\\
\midrule
Temporary cut, elasticity on price/regular & $-2.5$\\
Regular price, elasticity on regular/list & $-0.5$\\
Promotion & $+40\%$\\
Display & $+25\%$\\
Promotion and display together, extra & $+15\%$\\
Week after a planned week with nothing planned & $-15\%$\\
\midrule
Chance of a cut; depth & 12\%; 10\% to 40\%\\
Promotion chance, with a cut; without & 60\%; 3\%\\
Chance of a display & 8\%\\
\bottomrule
\end{tabular}
\end{table}

\subsection{What the checks catch}

\begin{figure*}[t]
\centering
\includegraphics[width=\textwidth]{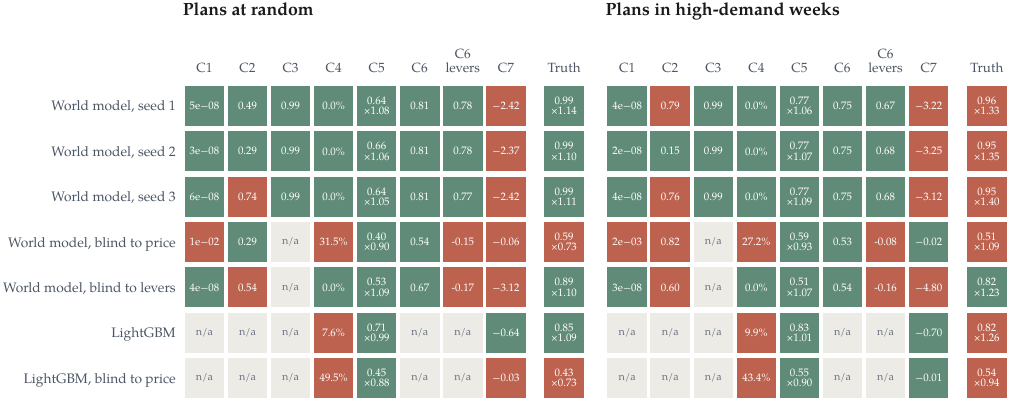}
\caption{\textbf{Correct models pass the checks; so do models whose
explanations are wrong in ways only some checks can see.} Each check's score
and its pass (green) or fail (red), for every
model in both worlds; grey, does not apply. C5: rank of predicted with observed
lift, and summed forecast over summed sales. C6 levers: the lowest per-lever
rank agreement with LightGBM, reported without a pass rule.
C7: pooled slope of the stated response to the price spells at the origin,
green when inside the 95\% interval of the measured slope, a rule added after
the robustness results (Appendix~\ref{app:plan}). Truth: rank
agreement of the model's lifts with the true lifts and the ratio of their sums,
green when the rank is at least 0.7 and the ratio within 0.8 to 1.25.}
\label{fig:knownchecks}
\end{figure*}

\begin{figure*}[t]
\centering
\includegraphics[width=\textwidth]{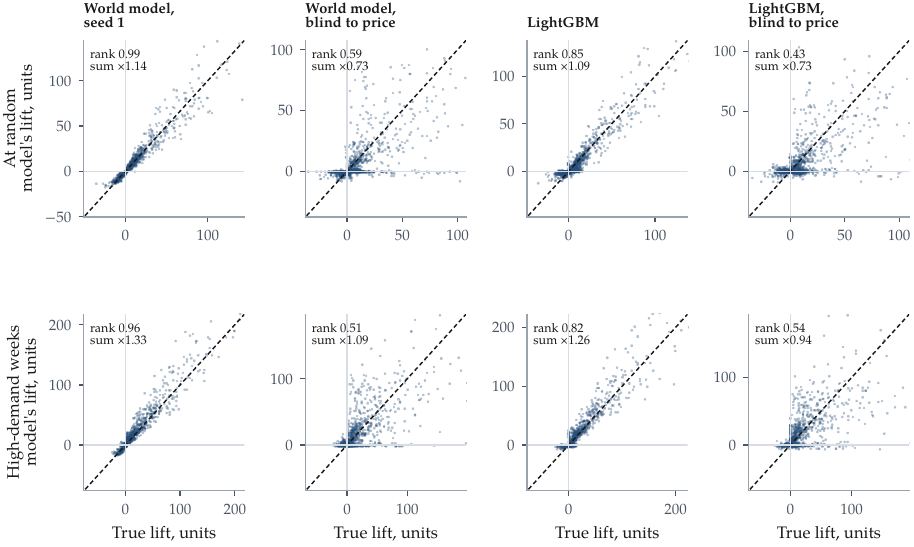}
\caption{\textbf{When plans follow demand, both the world model and LightGBM
overstate what the plan adds, and agree with each other.} Each point is an
item-week with something planned: the model's total lift against the true lift,
units. Dashed: equal. Top, plans at random; bottom, plans made more often in
high-demand weeks.}
\label{fig:knownscatter}
\end{figure*}

Figure~\ref{fig:knownchecks} shows every check on every model, and
Table~\ref{tab:known} in the appendix every score. Six results stand out.

\paragraph{Correct models pass, and their explanations are right}
With plans at random, the three world-model seeds pass every check that applies
except C7 and, for one seed, C2, and their lifts match the truth: rank agreement
\kaClWmTruth{} and summed lift \kaClWmBias{} times the true lift for seed 1,
with the price, promotion and display parts each within 13\% of the truth
(Table~\ref{tab:known}).
Seeds agree on their lifts as closely as on their forecasts (C3, \kaClSeedRank).
LightGBM's lifts rank \kaClLgbTruth{} with the truth at \kaClLgbBias{} times.

\paragraph{The price check catches a model blind to price, and flags LightGBM
too; the check against sales does not}
All four models fitted without a usable price fail C4: \kaClWmNpCfour\% and
\kaCoWmNpCfour\% of item-weeks for the world model, \kaClLgbNpCfour\% and
\kaCoLgbNpCfour\% for LightGBM, against 0\% for the world models fitted on real prices. It also fails both
LightGBM models fitted on real prices (\kaClLgbCfour\% and \kaCoLgbCfour\%),
whose lifts are close to the truth and whose regular-price response passes C7:
trees respond to price in steps, so a deeper cut sometimes lowers the forecast.
C4 is sensitive to a model ignoring price but not specific to it. All four pass C5: their forecasts on planned weeks sum to 0.88 to 0.93
of sales, with rank agreement 0.40 to 0.59. Their explanations are wrong in a
way a merchant would act on: the price part is zero, and the promotion, which
usually comes with a cut, takes its place, at 1.95 to 2.37 times its true lift
across the four (Table~\ref{tab:known}; Figure~\ref{fig:knownscatter}).
Both price-blind world models also fail C1, because the audit takes the
regular price from the true history while these models learned it from a
shuffled one; that is a property of how the failure was planted, not a sign
that C1 detects it.

\paragraph{Only a per-lever comparison catches lifts given to the wrong lever}
The per-lever comparison was added after the first results, without a pass rule
fixed in advance, and is exploratory.
The world model fitted without usable promotion and display flags passes C1,
C4 and C5, and its total lift agrees with LightGBM's (C6, \kaClNlCsix{} with
plans at random and \kaCoNlCsix{} when plans follow demand) and ranks
\kaClWmNlTruth{} with the truth (plans at random). Its explanation gives
the promotion \kaClWmNlPromoBias{} and the display \kaClWmNlDisplayBias{} times
their true lift, and the price \kaClWmNlPriceBias{} times
(Table~\ref{tab:known}). Every check on the total lift except C2 passes it.
Its per-lever agreement with LightGBM does not: \kaClNlCsixPromo{} for the
promotion and \kaClNlCsixDisplay{} for the display, against
\kaClWmCsixPromo{} for the promotion in the correct world model (seed 1, both
with plans at random).

\paragraph{When plans follow demand, the lifts are too large and C1 and C3 to C6 still pass}
In the second world the three world-model seeds pass C1, C3, C4, C5 and C6
(two of them fail C2, and all three fail C7), and
their forecasts are more accurate than in the first (item-by-store RMSSE
\kaCoWmRmsse{} against \kaClWmRmsse{}): a plan made in high-demand weeks tells
the model about the demand. Their lifts are \kaCoWmBiasLo{} to \kaCoWmBiasHi{}
times the true lift, and the display's \kaCoWmDispLo{} to \kaCoWmDispHi{}
times (Table~\ref{tab:known}).
C5 cannot see this: sales in planned weeks are high for the same reason, and
the forecasts sum to \kaCoWmCfive{} times them. C6 cannot see it either:
LightGBM, fitted to the same data, overstates by \kaCoLgbBias{} times, and the
two agree at \kaCoWmCsix. No check that compares the explanation with the
model, with the data or with another model fitted to the same data can find a
bias that the data themselves carry.

\begin{figure*}[t]
\centering
\includegraphics[width=\textwidth]{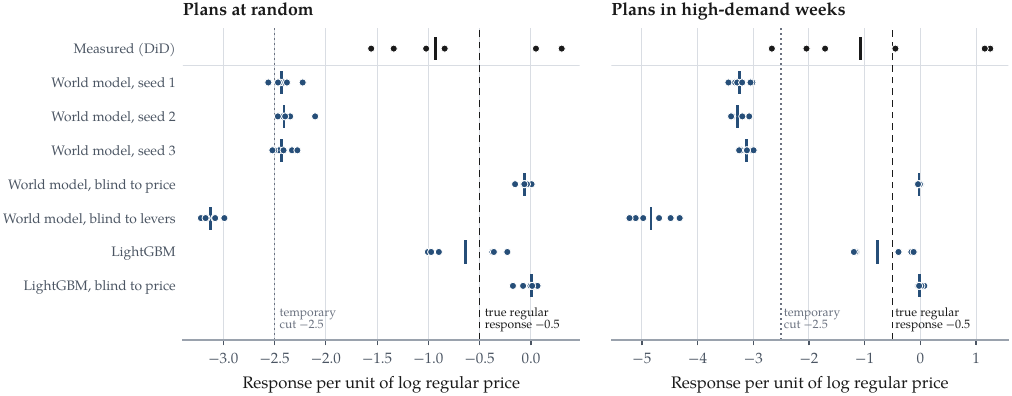}
\caption{\textbf{The world models fitted on real prices state a response to a
regular price change near or beyond the temporary-cut elasticity; LightGBM
states one near the truth, and the price-blind models almost none.} The stated change in
store category units per unit of log price for each of the six price spells
that start at the forecast origin (dots; bar, median), against the measured
response from difference in differences on the generated sales (top row), the
true response ($-0.5$, dashed) and the temporary-cut elasticity ($-2.5$,
dotted).}
\label{fig:knownc7}
\end{figure*}

\paragraph{The world model states a regular-price response near the temporary-cut elasticity}
For the price spells that start at the origin, the world model's stated
response is $\kaClWmCsevenLo$ to $\kaClWmCsevenHi$ per unit of log price across
seeds with plans at random and $\kaCoWmCsevenLo$ to $\kaCoWmCsevenHi$ when plans
follow demand, \kaClWmTimesLo{} to \kaCoWmTimesHi{} times the true $-0.5$,
and close to or beyond the temporary-cut elasticity of $-2.5$
(Figure~\ref{fig:knownc7}). Its response to a change in the regular price is of
the size of its response to a cut, not of the true regular-price response. LightGBM, whose inputs include the item's recent prices, states $\kaClLgbCsevenSlope$ and $\kaCoLgbCsevenSlope$. The measured
response from difference in differences on the generated sales is
$\kaClMeasSlope$ ($\kaClMeasLo$ to $\kaClMeasHi$) and $\kaCoMeasSlope$
($\kaCoMeasLo$ to $\kaCoMeasHi$): imprecise with six spells and three or four
controls each, as on Dominick's, but enough to reject the world model's answer
in both worlds and not LightGBM's. In the second world the measured interval
also contains zero, so there C7 passes the price-blind models too: with six
spells it can reject a response five times too large, not one that is absent. Apart from C2 for one seed, C7 is the only
check the world model fails in the first world, and the only one that finds this
error.

\paragraph{What the wrong lift costs a planner (added after the results)}
Two uses of the stated lift were scored against the truth in both worlds, from the saved lifts
(Appendix~\ref{app:plan}). Choosing which half of the planned item-weeks to keep by the
world model's stated lift captures \kdCoWmChoiceLo{} to \kdCoWmChoiceHi{} of the true lift the best
half carries when plans follow demand (\kdClWmChoiceLo{} to \kdClWmChoiceHi{} with plans at random;
LightGBM \kdCoLgbChoice{} and \kdClLgbChoice; a random half \kdCoRandom{} and \kdClRandom). The world
model ranks the weeks right. Ordering stock to its stated lift is another matter: the stated and true
lifts differ by \kdCoWmStockLo{} to \kdCoWmStockHi{} of the true lift when plans follow demand, against
\kdClWmStockLo{} to \kdClWmStockHi{} with plans at random (LightGBM \kdCoLgbStock{} and
\kdClLgbStock). An overstated lift chooses the right weeks and orders too much for them.

\paragraph{The overstatement holds when the world changes (added after the results)}
The world where merchants plan into high-demand weeks was regenerated with one setting changed at a
time: how strongly plans follow demand, how long demand shocks last, and the regular-price and
temporary-cut elasticities (\ksN{} worlds, Table~\ref{tab:known-sens}). In every one the world
model's summed lift is \ksWmBiasLo{} to \ksWmBiasHi{} times the truth, its stated regular-price
response \ksWmRegLo{} to \ksWmRegHi{} times the true one, and it passes \ksWmPassedHi{} of the
\ksWmApplicable{} checks it can be scored on, as in the main study; its stock error is
\ksWmStockLo{} to \ksWmStockHi{} of the true lift. LightGBM's summed lift is \ksLgbBiasLo{} to
\ksLgbBiasHi{} times the truth and its regular-price response \ksLgbRegLo{} to \ksLgbRegHi{} times the
true one. The size of the overstatement moves with the settings; that the explanation passes the
checks and overstates the effect does not.

\begin{table*}[t]
\centering
\caption{\textbf{The world model's lift stays overstated when the generated world changes (added after the results;
Appendix~\ref{app:plan}).} The world where merchants plan into high-demand weeks, regenerated with one
setting changed at a time; world model seed 1 and LightGBM, fitted and audited as in the main study. Lift / truth:
summed stated lift over the true lift on planned item-weeks. Regular price: stated response to a lasting
regular-price change over the true one. Checks: C1, C2, C4, C5 and C6 passed of those that apply (C3 needs three
seeds). Stock error: absolute difference between stated and true lift, summed, over the true lift.}
\label{tab:known-sens}
\footnotesize\setlength{\tabcolsep}{4pt}
\begin{adjustbox}{max width=\textwidth}
\begin{tabular}{lrrrrrrrr}
\toprule
 & \multicolumn{4}{c}{World model} & \multicolumn{4}{c}{LightGBM} \\
\cmidrule(lr){2-5}\cmidrule(lr){6-9}
World & Lift / truth & Regular price & Checks & Stock error & Lift / truth & Regular price & Checks & Stock error \\
\midrule
As in the main study & 1.33 & 6.4 & 4/5 & 0.44 & 1.26 & 1.4 & 1/2 & 0.50 \\
Planning into demand, half as strong & 1.23 & 5.7 & 4/5 & 0.41 & 1.23 & 1.4 & 1/2 & 0.57 \\
Planning into demand, 1.5 times as strong & 1.23 & 6.3 & 4/5 & 0.34 & 1.17 & 1.8 & 1/2 & 0.38 \\
Demand shocks shorter (persistence 0.4) & 1.28 & 5.8 & 4/5 & 0.40 & 1.26 & 1.7 & 1/2 & 0.46 \\
Demand shocks longer (persistence 0.9) & 1.26 & 6.1 & 4/5 & 0.41 & 1.22 & 1.5 & 1/2 & 0.48 \\
Regular-price elasticity $-0.25$ & 1.29 & 12.5 & 4/5 & 0.43 & 1.26 & 3.0 & 1/2 & 0.51 \\
Regular-price elasticity $-1.0$ & 1.24 & 3.0 & 4/5 & 0.39 & 1.19 & 0.7 & 1/2 & 0.45 \\
Temporary-cut elasticity $-1.5$ & 1.42 & 4.4 & 4/5 & 0.52 & 1.35 & 0.9 & 1/2 & 0.55 \\
Temporary-cut elasticity $-3.5$ & 1.24 & 7.9 & 4/5 & 0.37 & 1.16 & 2.0 & 1/2 & 0.44 \\
\bottomrule
\end{tabular}
\end{adjustbox}
\end{table*}

\begin{figure}[t]
\centering
\includegraphics[width=\columnwidth]{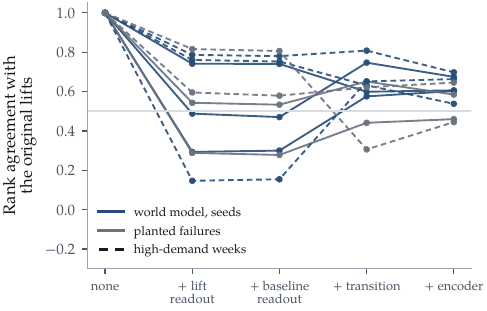}
\caption{\textbf{Re-initialising the network leaves the lifts' ranking largely
in place.} C2 for every world model, both worlds: rank agreement of the total
lift with the original after the lift readout, then the baseline readout, the
transition and the encoder are re-initialised in turn.}
\label{fig:knowncascade}
\end{figure}

\paragraph{Randomisation does not separate working models from broken ones}
After the lift readout is re-initialised, the rank agreement of the lifts with
the original is \kaLiftReadLo{} to \kaLiftReadHi{} across world models, and with
every part re-initialised it is \kaEncLo{} to \kaEncHi{}
(Figure~\ref{fig:knowncascade}). The construction itself ties the lifts to the
weeks with something planned and to the size of the item, so a network with
random weights still ranks them much as the trained one does. C2 fails correct
seeds and passes broken models; as specified, it does not test a lift built
into the model, and we do not use it as evidence below.

\paragraph{What the study says about the checks}
In an analysis added after the results, across the 14 models, the share of
checks passed tracks how well a model ranks
the true lifts (Spearman \kaPassTruthRho) but not how close its lifts are in
size (\kaPassBiasRho{} with the negative absolute log ratio): the models that
pass the most checks include the world models fitted where plans follow
demand, whose lifts are \kaCoWmBiasLo{} to \kaCoWmBiasHi{} times the truth. Accuracy tracks the gross failures, the price-blind models being the
least accurate, and misses the subtle ones. The checks earn their place as
follows. C4 finds a model that ignores price, and also flags one whose price
response is merely uneven. Per-lever agreement with another
model finds lifts given to the wrong lever. C1 and C3 confirm construction and
stability. C5 and C6 confirm agreement but cannot find a bias the data carry,
and C2 as specified finds nothing. Only a measured cause, C7, finds that a
model's stated response to a price change is wrong, and only one kind of
response, a sustained regular-price change, has a measured cause on public
retail data.

\section{Data, model versions and protocol}
\label{sec:versions}

\paragraph{Data}
The four public data sets of the \companion{} ~\cite{rwm}. \emph{M5} holds
daily Walmart sales of 3,049 items in 10 stores from 2011 to 2016, with prices,
event days and SNAP days~\cite{m5accuracy}. \emph{Dominick's Finer Foods} holds
weekly scanner data by store for a Chicago chain from 1989 to 1997, with price,
a promotion code and cost~\cite{dominicks}; the \companion{} 's six
categories are audited one by one (cheese, snack crackers, soft drinks, bottled
juice, analgesics, frozen entrees). \emph{Breakfast at the Frat} holds weekly
sales of 55 breakfast items in 77 stores from 2009 to 2012, with display,
feature and tag-only price reduction flags~\cite{frat}. \emph{Corporaci\'on
Favorita} holds daily sales of an Ecuadorian grocery chain to August
2017~\cite{favorita}: the 10 stores with the most units in 2016 crossed with the
items sold in all ten, with the promotion flag, holidays, events and paydays,
and no prices. That makes nine data sets. No raw data is redistributed.

\paragraph{Models audited}
The audit reads the final tuned world model the \companion{}  reports for
each data set: settings tuned by a search on that data set, fitted on its
rolling windows (six on each Dominick's category, four on Frat, M5 and
Favorita), with the fitted model of every window saved. Each is the average of
five copies fitted with different seeds, except on M5, where the final model
is one copy. C1 to C6 and C8 read these saved models, the runs whose scores
the \companion{}  reports; none was refitted or tuned for the audit. Two
checks need other fits with the same settings. C6 compares with each data
set's tuned LightGBM, refitted on the same windows. C7 needs models whose
training ends the week before each zone episode, so for it each Dominick's
category's final world-model and LightGBM settings are fitted again up to that
week. The question of whether passing the checks goes with accuracy uses the
versions the settings searches saved: every change tried on every data set,
one seed when screened and three when confirmed, each with a fitted model per
search window and its search-window accuracy. An earlier audit of the
\companion{} 's development versions, made before its models
were tuned, is reported in \S\ref{sec:results} as a record and is not the
paper's main result.

\paragraph{Protocol}
The audit reads models and never tunes them, and no audit result changed a
model setting. Every check runs on every version it applies to and every result
is reported, including failures. The plan, its thresholds and every later
entry are dated in \texttt{docs/PAPER2\_PLAN.md} and listed in
Appendix~\ref{app:plan}, with which entries were made before the results they
affect and which after; everything added after the results is labelled so
where it is reported.

\section{An observational benchmark from zone price deviations}
\label{sec:zones}

\subsection{Finding the deviations}

Dominick's set regular prices by zone: within a zone every store was meant to
carry the same regular price~\cite{dominicks}. For each store and item we take
the log of the week's regular price (no deal code, not more than 5\% below the
item's highest price in that store over the last 12 weeks) over that item's
median regular price in that store over all weeks, and subtract the median of
the same across the zone's stores that week. Moves made zone-wide cancel, and
so do an item's usual price differences between stores; a store moved on its
own stands out. An episode is at least 12 consecutive weeks in which this
quantity is at least 3\% in the same direction for at least 60\% of the store's
items, with a median of at least 5\% over the episode. Controls are stores of
the same zone within 1\% of their zone over those weeks. An episode with fewer
than three controls, or without the 26 comparison weeks before its start, is
listed but not scored.

The detector was written, and its episodes listed, before any model was scored
against them. A first version measured stores against the zone's most common
price and found one episode, in analgesics: in the experiments a third of a
zone's stores moved up and a third down, so the most common price was often a
moved one. The change is recorded, with its date, in the plan.

\subsection{Measuring their effect}

For episode $e$ of store $s$ with controls $\mathcal C_e$, the response is the
difference in differences on log category units,
\begin{equation}
  \log Y_{jw} = \alpha_j + \gamma_w + \tau_e\, D_{jw} + \varepsilon_{jw},
  \qquad j \in \{s\} \cup \mathcal C_e,
\end{equation}
over the 26 weeks before the episode and its first eight weeks, where $Y_{jw}$
is store $j$'s category units in week $w$ and $D_{jw}$ marks the treated store
in the episode weeks. Standard errors are clustered by
store~\cite{bertrand2004,cameron2015} with the CR1 correction and $t$ critical
values on stores minus one degrees of freedom, with \zStoresLo{} to
\zStoresHi{} stores per episode. The plan scores each episode by its own
interval and reports the pooled slope of $\tau_e$ on $\delta_e$ across
episodes. After the placebo showed the per-episode intervals to be too narrow,
an interval for the pooled slope robust to unequal variances was added, and the
pooled slope became the summary used for C7; both additions are exploratory. The estimate over
the full episode is kept in the list as well. The models forecast at most eight
weeks ahead, so each model is fitted on the data up to the week before an
episode and scored on its first eight weeks, the same weeks as $\tau_e$.

\subsection{What the deviations show}

\begin{figure*}[t]
\centering
\includegraphics[width=\textwidth]{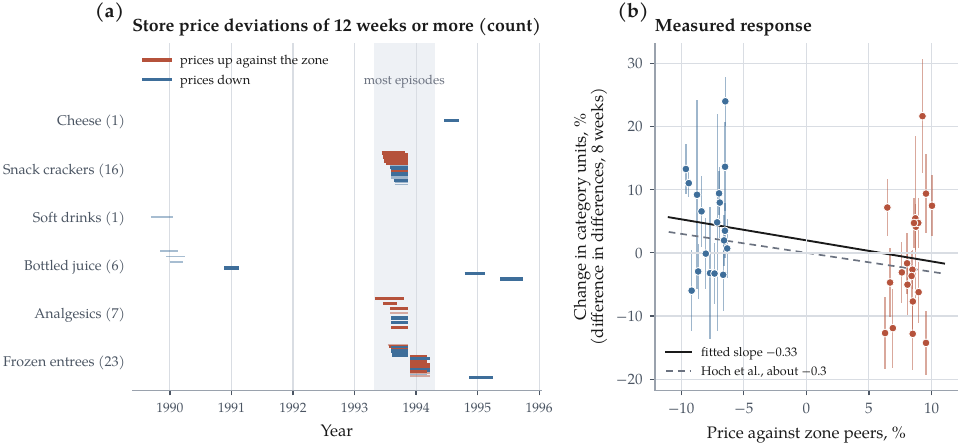}
\caption{\textbf{The price deviations cluster between spring 1993 and spring
1994, and on average sales move against price.}
\textbf{(a)} Every episode found in the six categories (count in brackets);
faint: fewer than three control stores, listed but not scored. \textbf{(b)} For
the \zScored{} scored episodes, the store's price against its zone peers and
the measured change in its category units over the episode's first 8 weeks
(difference in differences against same-zone stores, store and week fixed
effects, 95\% interval clustered by store; these intervals are too narrow, see
Figure~\ref{fig:zonechecks}). Solid line: fitted slope; dashed: the elasticity near $-0.3$ implied by the
experiments~\cite{hoch1994}.}
\label{fig:zones}
\end{figure*}

The detector finds \zEpisodes{} episodes in the six categories, \zScored{} with at
least three control stores (Figure~\ref{fig:zones}). Most lie between Dominick's
weeks 190 and 240, April 1993 to April 1994. Hoch, Dr\`eze and Purk published
their experiments in 1994, and the public files do not date them, so whether
these spells are those experiments cannot be checked. The scored stores' prices
sit \zSizeLo\% to \zSizeHi\% off their zone peers.
Over the first eight weeks, category units move against price in
\zSignShare\% of the \zScored{} scored episodes. The pooled slope of the measured response on
the price change is $\zSlope$, with a 95\% interval of $\zSlopeLo$ to
$\zSlopeHi$ (heteroskedasticity-robust across episodes), and over the full
episode $\zFullSlope$ ($\zFullLo$ to $\zFullHi$). Both are in line with the
elasticity near $-0.3$ implied by Hoch, Dr\`eze and Purk's report of a 3\% volume
change for a 10\% price change, and neither is precise. These intervals across
episodes were added after the placebo was seen (below).

\begin{figure*}[t]
\centering
\includegraphics[width=\textwidth]{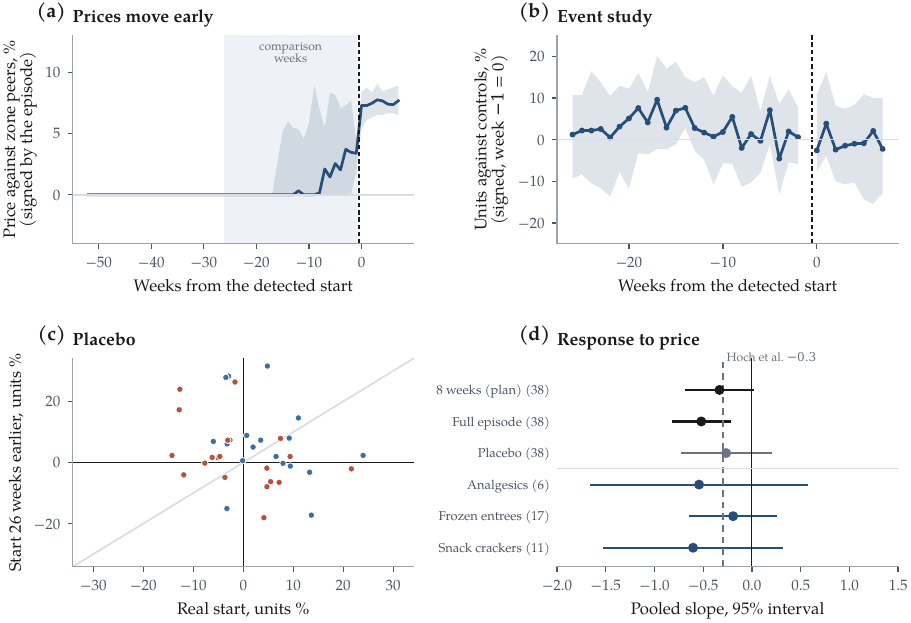}
\caption{\textbf{Prices start moving before the detected start, and
single-episode intervals are too narrow, so the pooled response is the one
compared with the models.}
\textbf{(a)} The store's regular prices against its zone peers, signed by the
episode's direction: median and interquartile band across the \zScored{} scored
episodes; shaded, the 26 comparison weeks of the difference in differences.
\textbf{(b)} Event study: weekly coefficients signed the same way, week $-1$ as
reference; median and interquartile band. \textbf{(c)} The 8-week estimate with
the start moved 26 weeks earlier, where prices had not moved, against the real
estimate; red, prices up; blue, down. \textbf{(d)} Pooled slopes of the
response on the price change with 95\% intervals, for the plan's 8-week
estimate, the full episode, the placebo, and each category with at least five
episodes (count in brackets). The detection rule, controls and estimator were
fixed in advance; panels (a) and (d) were added after the placebo was seen.}
\label{fig:zonechecks}
\end{figure*}

Three robustness analyses, fixed in advance, and two added after seeing them
qualify this benchmark (Figure~\ref{fig:zonechecks}).

\paragraph{Prices move before the detected start}
The detector dates an episode from the first week in which most of the store's
items are at least 3\% off. Prices ease in before that: in the eight weeks before
the detected start, \zEarlyShare\% of store-weeks are already 3\% or more off in
the episode's direction (panel a). Part of the 26 comparison weeks is therefore
treated, which biases the 8-week estimate towards zero and shows in the event
study as a gradual decline in signed sales over the weeks before the start
(panel b). The full-episode estimate, which compares the whole spell, is
larger, $\zFullSlope$ against $\zSlope$.

\paragraph{Single-episode intervals are too narrow}
Moved 26 weeks earlier, to weeks where prices had not moved, the same estimator
gives a pooled slope of $\zPlaceboSlope$ ($\zPlaceboLo$ to $\zPlaceboHi$), and
\zPlaceboSig\% of the \zPlaceboN{} placebo intervals exclude zero, where 5\%
should. The intervals clustered by store are far too narrow: their median
standard error is \zEpisodeSE{} while placebo estimates spread with a standard
deviation of \zPlaceboSD. With few stores per episode, weekly category sales that
move together over many weeks, and errors clustered by so few
stores, this is the failure Bertrand, Duflo and Mullainathan~\cite{bertrand2004}
and Cameron and Miller~\cite{cameron2015} describe. A single episode's
interval, and with it the plan's per-episode score for C7, cannot be trusted;
the comparison of models with the measured response is made on the pooled
slope and its interval.

\paragraph{The pooled response is stable but imprecise}
Over the nine detector settings (Table~\ref{tab:sensitivity}) the number of
scored episodes ranges from \zSensLo{} to \zSensHi{} and the pooled slope from
$\zSensSlopeLo$ to $\zSensSlopeHi$. By category, the slope is negative in
each of the \zCatN{} categories with at least five episodes, and every
category's interval is wide (panel d).

\begin{table}[t]
\centering
\caption{\textbf{The pooled response keeps its sign under every detector
setting.} Episodes found and scored, and the pooled 8-week slope, for each
weekly size and episode median; the plan's setting in bold. The settings and
the pooled slope were fixed in advance.}
\label{tab:sensitivity}
\footnotesize
\begin{tabular}{@{}rrrrr@{}}
\toprule
Each week & Episode median & Episodes & Scored & Slope \\
\midrule
2\% & 4\% & 64 & 45 & $-0.38$ \\
2\% & 5\% & 55 & 38 & $-0.37$ \\
2\% & 7\% & 38 & 26 & $-0.20$ \\
3\% & 4\% & 63 & 45 & $-0.32$ \\
\textbf{3\%} & \textbf{5\%} & \textbf{54} & \textbf{38} & \textbf{$-0.33$} \\
3\% & 7\% & 37 & 26 & $-0.16$ \\
4\% & 4\% & 59 & 43 & $-0.32$ \\
4\% & 5\% & 54 & 38 & $-0.34$ \\
4\% & 7\% & 37 & 26 & $-0.17$ \\
\bottomrule
\end{tabular}

\end{table}

What survives is a pooled response of store category units to a sustained
change in regular prices of $\zSlope$ over the first eight weeks, whose interval
just includes zero and overlaps the placebo's, and $\zFullSlope$ over the full
spell, whose interval excludes zero. The 8-week figure is pulled towards zero by
prices that moved before the detected start. With the placebo this close, these data cannot by themselves establish
that C7 separates a model that treats a regular price change like a temporary
cut, an elasticity several times larger, from one that does not; the
known-answer study establishes that (\S\ref{sec:known}). On public data the
comparison with the models is descriptive.

\paragraph{Limits of the benchmark}
It is not a ground truth. Which stores and weeks were treated is inferred from
prices, not recorded; prices move before the detected start; and the placebo at
false start dates gives a pooled slope ($\zPlaceboSlope$) close to the 8-week
estimate ($\zSlope$), so the 8-week estimate cannot be told from what the
estimator finds where nothing happened. The full-spell estimate
($\zFullSlope$, $\zFullLo$ to $\zFullHi$) has an interval that excludes zero,
but it too overlaps the placebo's ($\zPlaceboLo$ to $\zPlaceboHi$). Some deviations can be zone moves that were not
experiments, and the models were trained on data that include earlier
deviations. What the benchmark measures, noisily, is the response of a store's category units to a sustained change in its
regular prices relative to its zone, over eight weeks. It is not the response
to a temporary cut, which the \companion{} 's models mostly learn from, and a
model can be right about one and wrong about the other. The known-answer study
(\S\ref{sec:known}) has both, with different true elasticities, for that
reason.

\section{The audit of the final tuned world models}
\label{sec:tuned}

\textbf{The tuned models' explanations add up under each model's own plan for
nothing planned and carry the plan in their state, but on most data sets a
deeper price cut can lower the forecast, and both the world model and LightGBM
state a larger price response than the one measured.}

\begin{figure*}[t]\centering
\includegraphics[width=\textwidth]{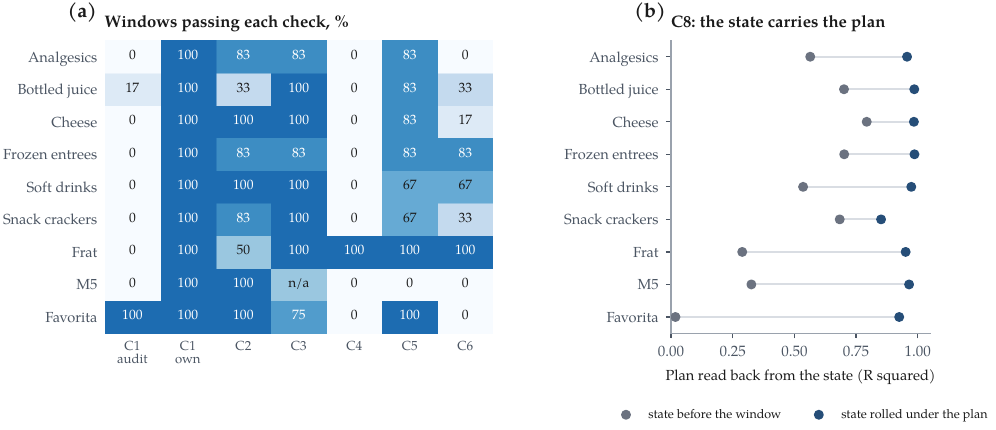}
\caption{\textbf{Final tuned world models: the state rolled through the plan
carries the plan on every data set.} \textbf{(a)} Share of each data set's
windows passing each check (Table~\ref{tab:tunedpass}). \textbf{(b)} C8 on
each model's first window: $R^2$ of a ridge probe reading the plan inputs of a
period back from the state before the window and from the state rolled through
the plan, averaged over the inputs (on Favorita, which has no prices, the
promotion, event and payday flags).}
\label{fig:tunedchecks}
\end{figure*}

The audit read the world model the \companion{} ~\cite{rwm} reports for each
of its \tSets{} data sets (six Dominick's categories, Frat, M5 and Favorita),
with one saved model per rolling window, \tWindows{} windows in all
(\S\ref{sec:versions}). The checks and their thresholds are those fixed before
the first audit (Table~\ref{tab:thresholds}); Table~\ref{tab:tunedpass} and
Figure~\ref{fig:tunedchecks} give the results. Pass rates below are shares of
all \tWindows{} windows pooled across data sets, with the count; the table
gives them by data set.

\begin{table*}[t]\centering
\caption{\textbf{The explanations add up under the model's own plan on every
window; C4 fails on every data set but Frat.} Share of each data set's windows
passing each check, \%, for its final tuned world model. C1 is scored under
the audit's plan for nothing planned (fixed in advance) and under the model's
own; scoring under the model's own plan was defined on 18 October, after the
earlier audit's results, and fixed for this audit on 13 November, before it
ran (Appendix~\ref{app:plan}). C2 is reported but not used as evidence
(\S\ref{sec:known}). C3 is the rank agreement of lifts between the model's five
copies (n/a on M5, whose final model is one copy); C6 the rank agreement of
total lift with the data set's tuned LightGBM. Thresholds in
Table~\ref{tab:thresholds}.}
\label{tab:tunedpass}
\footnotesize\setlength{\tabcolsep}{6pt}\begin{tabular}{@{}lrrrrrrrr@{}}
\toprule
Data & Windows & C1 (audit plan) & C1 (own plan) & C2 & C3 & C4 & C5 & C6 \\
\midrule
Analgesics & 6 & 0 & 100 & 83 & 83 & 0 & 83 & 0 \\
Bottled juice & 6 & 17 & 100 & 33 & 100 & 0 & 83 & 33 \\
Cheese & 6 & 0 & 100 & 100 & 100 & 0 & 83 & 17 \\
Frozen entrees & 6 & 0 & 100 & 83 & 83 & 0 & 83 & 83 \\
Soft drinks & 6 & 0 & 100 & 100 & 100 & 0 & 67 & 67 \\
Snack crackers & 6 & 0 & 100 & 83 & 100 & 0 & 67 & 33 \\
Frat & 4 & 0 & 100 & 50 & 100 & 100 & 100 & 100 \\
M5 & 4 & 0 & 100 & 100 & n/a & 0 & 0 & 0 \\
Favorita & 4 & 100 & 100 & 100 & 75 & 0 & 100 & 0 \\
\bottomrule
\end{tabular}
\end{table*}

\paragraph{C1: the parts add up under the model's own plan}
Under the model's own plan for nothing planned, every window of all \tSets{}
data sets passes C1. Under the audit's plan, \tCOneWin{} windows pass
(\tCOneWinPct\%). The two scorings differ only in the plan for nothing
planned, so the failures come from the difference between the audit's regular
price and the model's own, not from how the parts add up. Which series differ
was not traced for these models; for the earlier versions it was
(\S\ref{sec:results}).

\paragraph{C2 and C3: randomisation and agreement between copies}
After the lift readout is re-initialised, the lifts' rank agreement with the
original falls below 0.5 on \tCTwoWin{} windows (\tCTwoWinPct\%); as the
known-answer study showed, this does not test a lift built into the model and
is not used as evidence. The five copies of each model rank item lifts alike:
the mean rank agreement between copies, averaged over each data set's windows,
ranges from \tCThreeLo{} to \tCThreeHi{} across data sets, and \tCThreeWin{}
windows pass C3 (\tCThreeWinPct\%).

\paragraph{C4: on most data sets a deeper cut can lower the forecast}
C4 fails on every data set but Frat. A deeper price cut lowers the forecast in
\tCFourDomLo\% to \tCFourDomHi\% of item-week steps on the Dominick's
categories, \tCFourFav\% on Favorita and \tCFourMfive\% on M5, against
\tCFourFrat\% on Frat (each a mean over the data set's windows). Favorita records no prices and its price
input is held at 1, so a cut there is a value the model never saw in training,
and its C4 score does not measure a learned price response. The forecasts
are accurate on these data sets~\cite{rwm}, so this is a fault in how the model
responds to the depth of a cut, not in its forecasts of the plans it saw.
The failures are not cuts beyond the prices the model was trained on (added after the results,
Appendix~\ref{app:plan}). Counting only steps whose deeper price is no lower than the lowest
price the item had in that store in training (\cfoKeptDomLo\% to \cfoKeptDomHi\% of steps on Dominick's,
\cfoKeptMfive\% on M5), a deeper cut lowers the forecast in \cfoObsDomLo\% to \cfoObsDomHi\% of steps on
the Dominick's categories, against \cfoAllDomLo\% to \cfoAllDomHi\% on all steps, and \cfoObsMfive\% on
M5 against \cfoAllMfive\%; on Favorita, whose price is always 1, no step stays inside the range.

\paragraph{C5 and C6: agreement with sales and with LightGBM}
The forecasts in planned weeks match sales in level and order on \tCFiveWin{}
windows (\tCFiveWinPct\%, C5); M5 fails, ranking planned weeks with a
correlation of \tCFiveRankMfive{} against the 0.3 required. The world model and
the tuned LightGBM agree on which items gain most from the plan with a mean
rank correlation, by data set, of \tCSixLo{} to \tCSixHi{}; \tCSixWin{}
windows reach the 0.5 required (\tCSixWinPct\%, C6).

\begin{figure*}[t]\centering
\includegraphics[width=\textwidth]{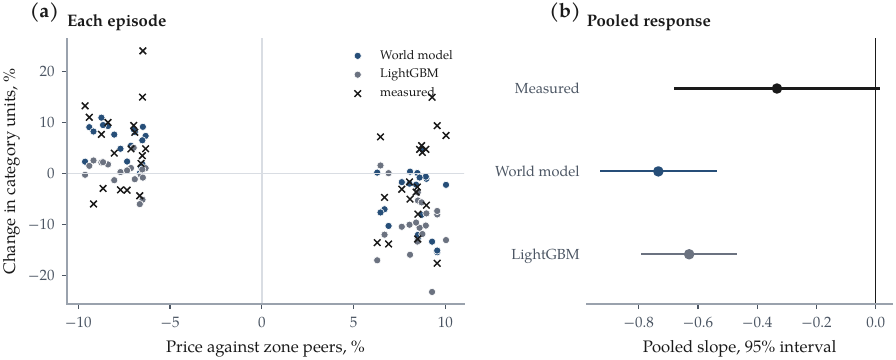}
\caption{\textbf{Final tuned models (this section): both state a larger price
response than the one measured.} \textbf{(a)} For each scored zone episode,
the change in category units the final tuned world model and LightGBM state for
the price deviation, and the change measured. \textbf{(b)} Pooled slopes with
95\% intervals (exploratory). The earlier audit's version of this comparison is
Figure~\ref{fig:realc7}.}
\label{fig:tunedc7}
\end{figure*}

\paragraph{C7: both models state a larger price response than the one measured}
On the \zScored{} scored zone episodes (\S\ref{sec:zones}), the measured pooled
slope of category units on price is $\tSlopeMeas$ (95\% interval
$\tSlopeMeasLo$ to $\tSlopeMeasHi$). The world model states $\tSlopeWm$
($\tSlopeWmLo$ to $\tSlopeWmHi$) and LightGBM $\tSlopeLgb$ ($\tSlopeLgbLo$ to
$\tSlopeLgbHi$) (Figure~\ref{fig:tunedc7}), each about twice the measured
response. This comparison on pooled slopes is exploratory
(\S\ref{sec:checks}, C7). The two models' intervals overlap each other and the
measured one, and the measured interval includes zero, so these episodes
exclude neither model's slope. In the known-answer study only the world model
overstated the regular-price response; LightGBM stated $\kaClLgbCsevenSlope$
and $\kaCoLgbCsevenSlope$ against a true $-0.5$. On Dominick's both overstate
it, and the measured response is too imprecise to say why.

\paragraph{C8: the state carries the plan}
On each model's first window, the plan inputs of a period can be read back from
the state rolled through the plan with a mean $R^2$, by data set, of
\tReadRolledLo{} to \tReadRolledHi{}, and rolling the state through the plan
adds \tReadGainLo{} to \tReadGainHi{} to what the state before the window holds
(Figure~\ref{fig:tunedchecks}b). The state keeps the plan it was given, and
the forecast and its lift are read from that state.

\paragraph{Checks against accuracy: no relation across the searches' versions}
The test was designed on 18 October, after the earlier audit's results, and
fixed for these versions on 13 November, before they were audited. Across the
\tSearchVersions{} model versions the settings searches saved
(\tConfigsAll{} configurations, each scored on the same data and windows as the
others of its search), the number of checks a configuration passed was not
related to its accuracy: the pooled rank correlation with item-by-store error
is $\tRhoAll$ (95\% interval $\tRhoAllLo$ to $\tRhoAllHi$, $p=\tPAll$), and
$\tRhoNoOneFive$ ($p=\tPNoOneFive$) leaving out C1 and C5. The interval
excludes all but a very weak relation in either direction.

\section{The earlier audit of the development versions}
\label{sec:results}

\begin{figure*}[t]\centering
\includegraphics[width=\textwidth]{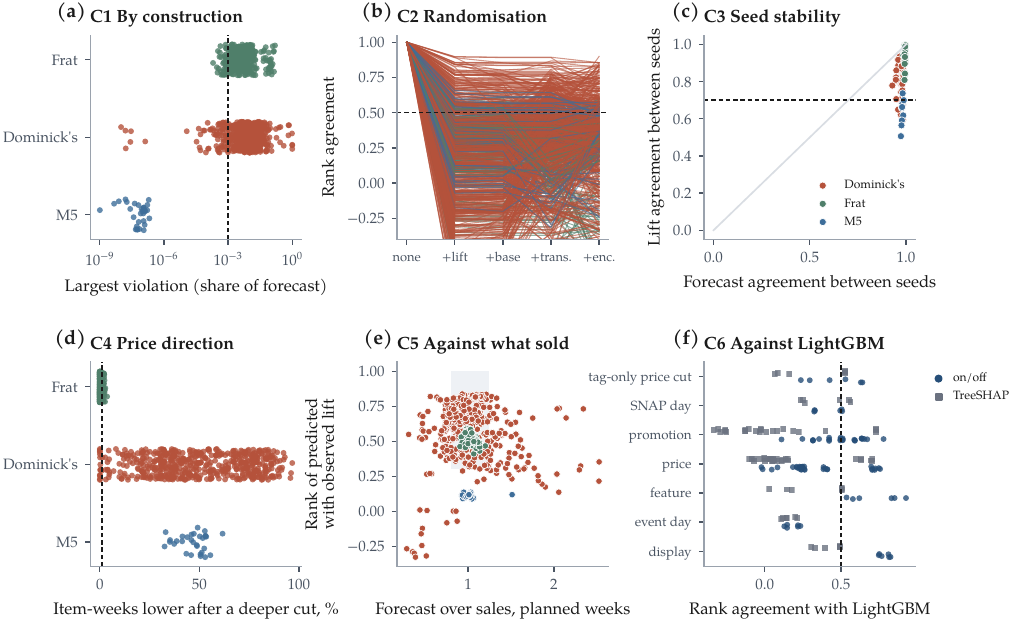}
\caption{\textbf{Earlier audit: the explanations agreed across
seeds on Dominick's and Frat, failed the price-direction check on Dominick's
and M5, and agreed with LightGBM only on Frat.} One point per version-window,
coloured by data set; dashed lines: pass thresholds. \textbf{(a)} C1, the
largest violation as a share of the largest forecast. \textbf{(b)} C2, rank
agreement of the total lift with the original as the lift readout, baseline
readout, transition and encoder are re-initialised in turn. \textbf{(c)} C3,
agreement of lifts between seeds against agreement of forecasts.
\textbf{(d)} C4, the share of item-week steps forecast lower after a deeper
price cut. \textbf{(e)} C5, forecast over sales and rank of predicted with
observed lift factor on planned item-weeks; shaded: the pass region.
\textbf{(f)} C6, rank agreement with LightGBM's on/off contributions (circles)
and TreeSHAP values (squares), by lever.}\label{fig:realchecks}
\end{figure*}

This section is the record of the earlier audit, made before the
\companion{} ~\cite{rwm} tuned its models. It is separate from the audit of
the final tuned models in \S\ref{sec:tuned} and is not the paper's main result.
It read every saved world-model and lift-model fit of the \companion{}  at
the time: \rVersions{} versions (a configuration and a seed), \rVerM{} on M5,
\rVerD{} on Dominick's and \rVerF{} on Frat, scored on \rWindows{}
version-windows. Most are development variants that differ in many settings at
once. No model was refitted or tuned. Table~\ref{tab:realpass} gives the pass
counts and Figure~\ref{fig:realchecks} every score.

\begin{table*}[t]\centering
\caption{\textbf{Earlier audit: no version passed every check on
M5 or Dominick's.} Pass counts by check and data set over version-windows,
under the rules fixed in advance (a check counted only where it applies: C3
needs two or more seeds, C6 a LightGBM run on the same window). C1 is scored
against the audit's plan for nothing planned; C1 under each model's own plan
was not scored for these versions. Thresholds in
Table~\ref{tab:thresholds}.}\label{tab:realpass}
\footnotesize\setlength{\tabcolsep}{6pt}\begin{tabular}{@{}lrrrrrrrrr@{}}
\toprule
Data & Versions & Windows & C1 & C2 & C3 & C4 & C5 & C6 & All \\
\midrule
M5 & 31 & 31 & 31/31 & 22/31 & 3/27 & 0/31 & 0/31 & 0/6 & 0/31 \\
Dominick's & 279 & 540 & 68/540 & 423/540 & 209/224 & 16/540 & 345/540 & 5/18 & 0/540 \\
Frat & 184 & 362 & 38/362 & 325/362 & 362/362 & 326/362 & 362/362 & 6/6 & 35/362 \\
\bottomrule
\end{tabular}
\end{table*}

\paragraph{Pass counts}
Scored as fixed in advance, \rAllF{} of \rWinF{} Frat version-windows pass
every check that applies, and none on M5 or Dominick's. On Dominick's and M5
the price-direction check (C4) and agreement with sales (C5) also fail, so
these counts do not rest on C1 alone.

\paragraph{C1: every failure sits where the two plans for nothing planned differ}
Baseline plus lift equals the forecast in every version (largest violation
\rCaPartsMax{} of the largest forecast), and the baseline moves with the plan by
at most \rCaBaseMax\% of it. The lift with nothing planned passes on all
\rVerM{} M5 versions but only on \rPassDCa{} Dominick's and \rPassFCa{} Frat
version-windows. The audit's plan sets each price to its regular price
(\S\ref{sec:checks}). Each model's own plan uses its own regular price over its
history window: the regular price the retailer recorded, where the
configuration reads one; or a \emph{held} price, one unchanged for a set number
of periods, where the configuration says so; otherwise the series' highest
price over its last 12 periods, falling back to its mean price over the window,
or 1.0 where the window records no price. A script added after the results
(\texttt{scripts/c1\_interface.py}) compared the two plans, which depend on the
data and settings but not on the fitted weights, for all \cIfOf{}
version-windows (M5 on three of its ten stores, for memory). They differ in
\cIfAnyD{} Dominick's, \cIfAnyF{} Frat and \cIfAnyM{} M5 version-windows. All
\cIfFailN{} version-windows that fail C1 have a difference, and all \cIfSameN{}
without one pass; since the plans differ in \cIfDiffN{} of the \cIfOf, this
alone is weak evidence. Of the series-periods that differ, \cIfShareNoReg\%
belong to series with no price in the audit's last 12 periods, where the model
falls back to its mean price or 1.0 (1.0 in \cIfShareOne\% of all differing
series-periods); \cIfShareRule\% to configurations that read a recorded or
held regular price; and \cIfShareNew\% to items not seen in training, whose
baseline borrows several other series' regular prices. C1 under each model's
own plan was measured only for the final tuned models (\S\ref{sec:tuned}),
where it passes on every window; it was not rescored for these versions. A
fault a merchant would care about remains: for an item with no recent price the
model's baseline is computed at a price nobody would plan.

\paragraph{C2 to C6}
C2 is reported but not used as evidence (\S\ref{sec:known}): the lifts' rank
agreement after the lift readout is re-initialised falls below 0.5 in
\rPassDCb{} Dominick's, \rPassFCb{} Frat and \rPassMCb{} M5 version-windows.
Seeds agree on the lifts with a median rank agreement of \rStabLiftF{} on Frat
and \rStabLiftD{} on Dominick's but \rStabLiftM{} on M5, against \rStabFcM{} to
\rStabFcF{} for the forecasts themselves; on M5, where prices rarely change and
the plan is mostly event and SNAP days, fits that agree on the forecast
disagree on how much of it the plan explains. A deeper cut lowers the forecast
by more than 0.1\% in a median of \rMedDCdPct\% of Dominick's and
\rMedMCdPct\% of M5 item-week steps, against \rMedFCdPct\% on Frat. A
diagnosis added after the results reran C4 on \dRuns{} saved runs
(\dWindows{} windows) chosen by a rule written before the rerun, which took
the runs with the largest C1 violations and others at random; their Dominick's
share is therefore higher than the full audit's (median \dCdAllD\% against
\rMedDCdPct\%). Cutting only a random tenth of series gave the same share
(\dCdTenthD\% on Dominick's, \dCdTenthM\% on M5, against \dCdAllD\% and
\dCdAllM\%), and the drops are large: the forecast falls by more than 5\% in a
median of \dCdBigD\% of Dominick's and \dCdBigM\% of M5 item-week steps. On
planned item-weeks the forecast is close to sales in total (median bias
\rMedDCeBias, \rMedFCeBias{} and \rMedMCeBias), and the rank agreement of
predicted with observed lift factor is \rMedDCeRank{} on Dominick's and
\rMedFCeRank{} on Frat but \rMedMCeRank{} on M5, so no M5 version passes C5.
Agreement with LightGBM's total lift passes on Frat (median \rMedFCf) and
mostly fails on Dominick's (\rMedDCf) and M5 (\rMedMCf).

\begin{figure*}[t]\centering
\includegraphics[width=\textwidth]{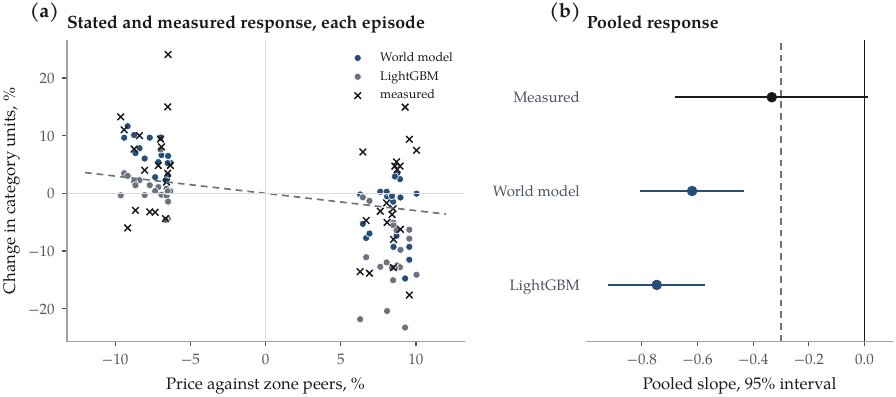}
\caption{\textbf{Earlier audit: the development world model and
LightGBM state about twice the measured response to a regular price change.}
\textbf{(a)} For each scored Dominick's zone episode, the store's price against
its zone peers and the change in its category units over the first eight weeks:
stated by the world model and LightGBM, each fitted up to the week before the
episode, and measured by difference in differences; dashed: an elasticity of
$-0.3$. \textbf{(b)} Pooled slope across episodes with a 95\% interval robust
to unequal variances (exploratory). The tuned audit's version is
Figure~\ref{fig:tunedc7}.}
\label{fig:realc7}
\end{figure*}

\paragraph{C7}
Over the \rZEpisodes{} scored zone episodes, the pooled slope of the stated
response on the price deviation (exploratory) is \rZSlopeWorld{} for the
world model (95\% interval \rZSlopeWorldLo{} to \rZSlopeWorldHi{}) and
\rZSlopeLgb{} for LightGBM (\rZSlopeLgbLo{} to \rZSlopeLgbHi{}), against
\rZSlopeMeas{} (\rZSlopeMeasLo{} to \rZSlopeMeasHi{}) measured
(Figure~\ref{fig:realc7}). Both are about twice the measured response, and the
measured interval overlaps both. LightGBM's slope is the larger here and the
world model's in the tuned audit; the two models' intervals overlap in both
audits, so neither audit orders them. On the per-episode score fixed in
advance, the world model's stated response falls inside the measured interval
in \rZWorldInside{} of \rZEpisodes{} episodes and LightGBM's in \rZLgbInside,
but those intervals are too narrow to trust (\S\ref{sec:zones}). The stated
response has the expected sign in \rZWorldSign{} and \rZLgbSign{} episodes,
and the measured one in \rZMeasuredSign; this count re-estimates each episode
over the eight weeks the model forecasts, which start up to three weeks before
the detected start, and so differs by one from the episode list
(\S\ref{sec:zones}).

\paragraph{The main test}
On the unit fixed before any result, a version-window, the correlation between
the number of checks passed and item-by-store WRMSSE is \rRhoPassedM{} on M5,
\rRhoPassedD{} on Dominick's and \rRhoPassedF{} on Frat; pooled,
\rRhoPassedPool{} (permutation $p = \rPPassed$; Table~\ref{tab:realmain}).
Averaged over each version's windows it is \rRhoPassedPoolV{}
($p = \rPPassedV$). The answer moves with the unit, and neither permutation
allows for the dependence between versions that share a configuration or a
window. A test added after the results (18 October) holds the data and window
fixed: seeds averaged, Spearman correlation across configurations within each
set of configurations scored on the same series and window (\rWwData{} sets
with at least four configurations, \rWwConfigs{} configurations; M5 has none),
pooled weighted by the number of configurations, with a permutation and a
bootstrap that move whole configurations within each set. It treats
configurations that differ in one setting as exchangeable, so its interval and
$p$-value do not allow for that dependence. The pooled correlation is
\rWwRho{} (95\% interval \rWwLo{} to \rWwHi, $p = \rWwP$), \rWwNoCeRho{}
(\rWwNoCeLo{} to \rWwNoCeHi) with C5 left out, and \rWwNoCaCeRho{}
(\rWwNoCaCeLo{} to \rWwNoCaCeHi) with C1 and C5 left out. C5 compares the
forecast with sales over the same baseline and so shares information with
accuracy; on the unit fixed in advance it is the check most strongly related to
accuracy (rank \rRhoCeRankPool, bias \rRhoCeBiasPool). On these versions,
configurations that passed more checks did not clearly forecast better or
worse; the intervals exclude a strong relation in either direction, not a weak
one. The same test on the tuned searches' versions is in \S\ref{sec:tuned}.

\begin{table*}[t]\centering
\caption{\textbf{Earlier audit: across versions, the number of
checks passed was not clearly related to accuracy; the check that was, C5,
shares information with accuracy.} Spearman correlation between each score and
item-by-store WRMSSE (negative: higher scores go with lower error), within each
data set and pooled by permutation within data set (10,000 permutations). Left:
over version-windows, the unit fixed in advance. Right: over versions, the mean
over each version's windows. Both permutations treat their units as
independent. $^\dagger$Added after the results.}\label{tab:realmain}
\footnotesize\setlength{\tabcolsep}{6pt}\begin{tabular}{@{}lrrrrrrr@{}}
\toprule
 & \multicolumn{5}{c}{Version and window} & \multicolumn{2}{c}{Version} \\
\cmidrule(lr){2-6}\cmidrule(l){7-8}
Score & M5 & Dom. & Frat & Pooled & $p$ & Pooled & $p$ \\
\midrule
Checks passed & $-0.32$ & $-0.19$ & $0.26$ & $-0.08$ & 0.205 & $-0.19$ & 0.004 \\
C1 violation & $0.10$ & $-0.18$ & $-0.41$ & $-0.17$ & 0.007 & $-0.03$ & 0.664 \\
C2 agreement after randomising & $-0.11$ & $0.20$ & $-0.02$ & $0.02$ & 0.740 & $0.01$ & 0.842 \\
C3 seed agreement & $-0.07$ & $0.32$ & $0.30$ & $0.18$ & 0.010 & $-0.01$ & 0.938 \\
C4 share lower & $-0.06$ & $-0.04$ & $-0.18$ & $-0.09$ & 0.159 & $-0.01$ & 0.916 \\
C5 rank & $-0.19$ & $-0.32$ & $-0.75$ & $-0.42$ & 0.000 & $-0.32$ & 0.000 \\
C5 bias & $0.62$ & $0.03$ & $0.70$ & $0.45$ & 0.000 & $0.36$ & 0.000 \\
C6 agreement with LightGBM & $-0.31$ & $-0.48$ & $0.20$ & $-0.20$ & 0.394 & $-0.20$ & 0.392 \\
Checks passed, C5 left out$^\dagger$ & $-0.32$ & $-0.04$ & $0.26$ & $-0.03$ & 0.615 & $-0.13$ & 0.055 \\
Checks passed, C1 and C5 left out$^\dagger$ & $-0.32$ & $-0.10$ & $0.15$ & $-0.09$ & 0.163 & $-0.15$ & 0.032 \\
\bottomrule
\end{tabular}
\end{table*}

\section{Discussion}
\label{sec:discussion}

\paragraph{What the known-answer study shows}
The world model's explanation is faithful by construction: the baseline does not
move with the plan, the lift is zero when nothing is planned, and the two add up
to the forecast. In the known-answer study that explanation was right about what
each lever adds when plans were made at random, and wrong in two ways that no
check against the model could see. When merchants plan into strong weeks, the
lifts are \kaCoWmBiasLo{} to \kaCoWmBiasHi{} times the truth, because the data
tie the plan to demand the model cannot see. And asked about a change in the
regular price, the model states a response near the temporary-cut elasticity,
\kaClWmTimesLo{} to \kaCoWmTimesHi{} times the truth. Molecular attributions in
MolSanity~\cite{molsanity} came apart from the truth for a similar reason: a
model can be faithful to what it learned and what it learned can be wrong.

\paragraph{What this means for a merchant reading a lift}
In the known-answer study, a lift read from a model fitted to the store's own
history inherited its planning habits. If promotions go into weeks that would
sell well anyway, the model's promotion lift overstates what the promotion adds,
and an accurate forecast is no protection: the correct world-model seeds and
LightGBM forecast better when plans followed demand than when they were made at
random. The check against what sold, the one most often run in practice, agrees
with the overstated lift for the same reason. Two things follow. Agreement
between models fitted to the same data is not confirmation. And a model's
answer to a question it was not trained on, here a lasting change in the
regular price from a model that mostly saw temporary cuts, needs checking
against a measured response before it is used to set prices. The cost falls on quantities, not on
choices: an overstated lift still ranks the weeks to promote correctly, so a merchant using it to pick
promotions loses little, but one ordering stock to it buys for sales that do not come.

\paragraph{Which checks to run}
The known-answer study ranks the checks. The price-direction check (C4) is cheap
and caught every model blind to price, though it also flags LightGBM, whose
response to price comes in steps. Per-lever agreement with a second model, an
exploratory comparison, caught lifts given to the wrong lever, which every
check on the total lift missed. Construction (C1) and seed stability (C3) are
worth confirming once. The check against what sold (C5) and agreement with
LightGBM on the total lift (C6) confirm agreement, not correctness.
Randomisation (C2), the standard sanity check for
attributions~\cite{adebayo2018}, does not test an explanation built into the
model: re-initialising every part left the lifts' ranking largely in place.
Versions of C2 on the lift as a share of the baseline, or within planned weeks
only, were not tested. A measured cause (C7) is the only check that found the
regular-price error, though with six spells it could not tell a missing price
response from a correct one. No check in this study finds the bias from
planning into strong weeks; a measured promotion effect, which public retail
data do not offer, was not tested.

\paragraph{Measured causes on public data are scarce and noisy}
Dominick's zone price deviations give a pooled 8-week response close to a
placebo at false dates and a full-episode response in line with the chain's
own experiments; intervals clustered by a handful of stores are far too narrow,
and prices start moving before the detector sees them. A retailer running its
own tests would do better with the design under its control: stores assigned at
random, a known start date, and enough stores per arm for honest intervals.

\paragraph{The audits of the \companion{} 's models}
On the final tuned world models, the explanations add up under each model's own
plan for nothing planned on every window, the copies agree, and the state
carries the plan. Two results matter to a merchant. First, on Dominick's
and M5 a deeper cut can lower the forecast, a fault no accuracy score
revealed and one a merchant setting prices from the model would meet directly;
the earlier audit found the same on Dominick's and M5. Second, both the world
model and LightGBM state about twice the measured response to the zone price
deviations. In the known-answer study only the world model overstated this
response, so the real data do not reproduce that pattern, and the measured
interval is too wide to say which model is closer. The earlier audit also showed
that C1 under the audit's plan measures the distance between two definitions of
nothing planned, and that for an item with no recent price the model computes
its baseline at its mean price or at 1.0, which no merchant would call nothing
planned. An audit has to read the model's own definition of nothing planned,
and check it, before C1 means anything. Across the searches' versions, passing
more checks was not related to accuracy. Those versions have no known answer, so this result is about
accuracy only; whether an explanation is right is shown by the known-answer study, not by it. For a retailer this means two separate
tests: an accuracy comparison to choose the forecast, and checks of the
explanation, with a price-direction test and a measured cause among them,
before its lifts are used to plan.

\section{Conclusion}
\label{sec:conclusion}

Plan-lift explanations built into a forecasting model can be faithful, stable
and in agreement with sales and with another model, and still be wrong about
what a plan causes. In generated store data with known effects, the checks that
compare an explanation with the model, the data or another model caught a model
blind to price and, compared lever by lever, lifts given to the wrong lever.
They did not catch lifts inflated by planning into strong weeks, or a model
whose response to a regular price change is near its response to a temporary
cut. Only a measured cause found the second. On public retail data the one such
cause we could measure, Dominick's zone price deviations, gives a pooled
response near the chain's experiments over the full episode, but its 8-week
estimate is close to a placebo at false dates, so on public data the comparison
is descriptive. The final tuned world models of the \companion{}  add up
under their own plan for nothing planned on every window, but on most data sets
a deeper cut can lower their forecast, and both they and LightGBM state about
twice the measured response to a regular price change. Across the
configurations the settings searches tried, the number of checks passed was not
related to how well they forecast. On these models and data sets, an accurate
forecast was not evidence of a correct explanation, and the explanation has to
be tested on its own.

\section{Limitations and threats to validity}
\label{sec:limits}

\paragraph{The benchmark is observational and covers one effect on one data set}
C7 compares a model with a response estimated from observational data: the
treated stores and weeks are inferred, and the 8-week estimate is close to a
placebo at false dates. That C7 can tell a response several times too large from
a correct one is shown in the known-answer study, where the truth is known. On
Dominick's the benchmark is too noisy to establish even that, and comparing the
models with it is descriptive. It measures the response
of a store's category units to a sustained change in its regular prices. It says nothing directly about promotions,
displays or temporary cuts, or about M5, Frat and Favorita, where no such variation is used. A model can be right about regular prices and wrong about
promotions; the known-answer study shows how far the other checks go in
covering the rest.

\paragraph{The deviations are inferred}
The public files do not mark the experimental stores or weeks. Some deviations
can be zone moves made for other reasons, and a store moved for a reason that
also moves its sales would bias $\tau_e$. The event study and the placebo
address this for trends that start before the price moves; they cannot rule out
a shock that starts in the same week.

\paragraph{Few control stores}
Zones have few stores, so each scored episode has \zControlsLo{} to
\zControlsHi{} controls. Intervals clustered
by store with so few clusters are too narrow (\S\ref{sec:zones}); the pooled
slope across episodes carries the comparison.

\paragraph{The baseline is a choice}
Every lift is measured against nothing planned, with the audit's regular price
taken as the highest of the last 12 training periods. A different rule gives
different baselines and lifts. The audit applies its rule the same way to every
model and to the truth in the known-answer study; each model's own regular
price can differ from it.

\paragraph{The versions are not a designed sample}
The audited versions are what the \companion{}  saved. They share data and
differ in many settings at once, so the correlation across versions is
descriptive; it is computed within data set and tested by permutation within
data set, but versions are not independent draws, and the windows of one
version differ in difficulty, so the result depends on whether version-windows
or versions are the unit. Both are reported.

\paragraph{The audit's definition of nothing planned}
The audit builds ``nothing planned'' from the data: levers off, the price at
the series' highest price over the last 12 training periods, and the planned
price where the series has no price in those periods. The models build their
own from their history window: a recorded or held regular price where the
configuration says so, otherwise the highest price over the last 12 periods,
falling back to the series' mean price over the window, or to 1.0 where it
records no price. C1 under the audit's plan therefore measures the distance
between two definitions as much as the model; under the model's own plan it
passes on every window of the final tuned models (\S\ref{sec:tuned}) and was not
scored for the earlier versions (\S\ref{sec:results}). Neither
definition is the retailer's regular price, which these data do not record.

\paragraph{Refits for C6 and C7}
C1 to C5 and C8 read the saved final models. C6 compares them with LightGBM
refitted on the same windows, and C7 uses world models and LightGBM fitted
again with the final settings up to the week before each episode. These refits
are not the runs whose scores the \companion{}  reports, and the difference
between them and those runs was not measured.

\paragraph{The generated world is simpler than retail}
The known-answer data are multiplicative in the plan, with one shock process
and no cannibalisation between items. They test whether a check can catch a
failure, not how often the failure occurs in practice. The sensitivity worlds change one setting at a
time around one world, and the decision scores (choosing half the planned weeks, ordering to the stated
lift) are simple stand-ins for a planner's choices.

\section{Reproducibility, compute and licence}
\label{sec:repro}

Every number, table and figure in this paper is generated from committed
records by \texttt{make -C paper2}, and a continuous integration job rebuilds
the paper from a fresh checkout on every change and fails on any LaTeX error,
undefined reference or missing font. The checks are in
\texttt{src/rwm/audit/}; the zone detector and estimators in
\texttt{src/rwm/audit/zones.py}; the generated data in
\texttt{src/rwm/data/synthetic\_levers.py}. The runs:
\begin{itemize}\setlength{\itemsep}{1pt}
\item \nolinkurl{scripts/zone_episodes.py}: the episodes and their responses
(\nolinkurl{results/paper2/zone_episodes.csv});
\item \nolinkurl{scripts/zone_robustness.py}: the event study, placebo and
sensitivity;
\item \nolinkurl{scripts/known_answer.py}: the known-answer study;
\item \nolinkurl{scripts/audit_runs.py} and \nolinkurl{scripts/paper2_colab.sh}:
the audits, C6 and C7;
\item \nolinkurl{scripts/audit_diagnose.py}: the diagnosis of C1 and C4 in the
earlier audit (\nolinkurl{results/paper2/diagnose/});
\item \nolinkurl{scripts/c1_interface.py}: where the audit's plan for nothing
planned and each model's own differ, for every configuration and window of the
earlier audit (\nolinkurl{results/paper2/diagnose/c1_interface.csv});
\item \nolinkurl{scripts/audit_summary.py}: the tables committed under
\nolinkurl{results/paper2/}.
\end{itemize}

\paragraph{Compute}
The known-answer study fits 14 models on a two-core processor: each world model in 5 to 23 minutes (69 minutes in all, the longest while another job shared the processor), each LightGBM in seconds. The zone episode list, event study, placebo, sensitivity and price path read only the data and take minutes on the same processor. The audit of the saved versions runs on a Colab GPU; its run times are recorded with its results.

\paragraph{Copyright and licence}
\iftmlr Withheld for anonymised review.\else
Copyright \copyright{} 2026 Karthik Iyer. Code under the PolyForm Noncommercial
License~1.0.0; text, figures and tables under CC~BY-NC~4.0. Industry and
commercial use of the code, the manuscript or any part of them, including use
inside a company, is prohibited unless approved in writing by the author.\fi

\paragraph{Code}
\iftmlr The code, run records and result files are in the anonymised supplementary material; the public
repository and its archived release are withheld for review.\else
The code is at
\url{https://github.com/karthik-iyer-au/retail-world-models}; the version used
for this paper is archived as release v1.1.0 on Zenodo,
\href{https://doi.org/10.5281/zenodo.24271672}{doi:10.5281/zenodo.24271672}.\fi

\paragraph{Data availability}
M5 is available from Kaggle~\cite{m5accuracy}, Dominick's from the Kilts Center
for Marketing at the University of Chicago Booth School of
Business~\cite{dominicks}, and Breakfast at the Frat from
dunnhumby~\cite{frat}, and Corporaci\'on Favorita from
Kaggle~\cite{favorita}, each under its own terms; the repository fetches them and
redistributes none.

\section{Declaration of generative AI and AI-assisted technologies}
\label{sec:ai}
The research questions, scope, study design, choice of data and methods, and
interpretation of results are the author's own. Claude (Anthropic) was used as an
assistant to write and test code, run analyses, produce figures and tables, and
draft and edit text. The author directed this work throughout, reviewed every
result, and corrected and revised the code, analyses and text over many
iterations, and takes full responsibility for the content of the publication.

\onecolumn
\appendix
\section{Every zone price episode}
\label{app:episodes}

\begin{table}[H]
\centering
\caption{\textbf{All \zEpisodes{} episodes the detector finds, with their measured
response.} Weeks: Dominick's week numbers of the episode. Price: the store's
median log price against its zone peers over the episode, \%. Controls: same-zone
stores within 1\% of their zone. 8 weeks: difference-in-differences estimate on
log category units over the first eight weeks, \%, with its interval clustered
by store (too narrow, \S\ref{sec:zones}); Full: over the whole episode.
\textsuperscript{*} Listed, not scored: fewer than three controls, or
fewer than the 26 comparison weeks before the start (the four episodes starting
before week 27). An interval of zero width comes from an episode with one
control store, where the clustered standard error is zero, or, for soft drinks
store 105, from an episode with no comparison weeks.}
\label{tab:episodes}
\scriptsize\renewcommand{\arraystretch}{0.86}
\begin{tabular}{@{}llrrrrrrl@{}}
\toprule
Category & Zone & Store & Weeks & Price & Controls & 8 weeks & 95\% interval & Full \\
\midrule
Analgesics & 2 & 50 & 190--215 & 8.8 & 16 & 4.1 & 0.6 to 7.6 & −5.7 \\
Analgesics & 2 & 49 & 197--209 & 8.5 & 16 & −2.7 & −5.6 to 0.3 & −3.2 \\
Analgesics & 2 & 48 & 203--218 & 8.1 & 11 & −5.0 & −9.8 to −0.2 & −7.1 \\
Analgesics & 7 & 109 & 203--218 & 6.0 & 2 & 8.2 & −4.1 to 20.4 & 9.1 \textsuperscript{*} \\
Analgesics & 1 & 32 & 204--218 & −6.5 & 7 & 13.6 & 6.7 to 20.6 & 14.9 \\
Analgesics & 2 & 88 & 204--218 & −6.7 & 10 & −3.5 & −9.1 to 2.2 & −2.0 \\
Analgesics & 2 & 89 & 204--218 & 6.3 & 10 & −12.7 & −18.3 to −7.1 & −9.4 \\
Bottled juice & 12 & 105 & 9--24 & −7.9 & 7 & −7.2 & −18.9 to 4.4 & −3.2 \textsuperscript{*} \\
Bottled juice & 1 & 123 & 14--30 & −5.5 & 11 & 7.8 & 3.6 to 12.1 & 11.5 \textsuperscript{*} \\
Bottled juice & 6 & 40 & 17--28 & −6.6 & 6 & 8.2 & −9.0 to 25.3 & 5.8 \textsuperscript{*} \\
Bottled juice & 6 & 40 & 63--75 & −7.7 & 6 & −3.2 & −13.5 to 7.1 & −0.1 \\
Bottled juice & 2 & 84 & 266--283 & −9.4 & 24 & 11.0 & 8.9 to 13.2 & 12.7 \\
Bottled juice & 2 & 84 & 296--315 & −9.2 & 22 & −6.0 & −12.3 to 0.3 & −0.9 \\
Cheese & 6 & 77 & 249--261 & −7.1 & 7 & 4.8 & −12.3 to 21.9 & 4.1 \\
Frozen entrees & 5 & 8 & 201--218 & 10.2 & 1 & −26.8 & −26.8 to −26.8 & −31.0 \textsuperscript{*} \\
Frozen entrees & 2 & 47 & 202--218 & 6.7 & 9 & −4.7 & −10.1 to 0.7 & −3.2 \\
Frozen entrees & 2 & 28 & 204--218 & −6.9 & 9 & 8.0 & 2.4 to 13.5 & 7.9 \\
Frozen entrees & 1 & 71 & 204--218 & 9.3 & 4 & 21.6 & 12.7 to 30.6 & 17.1 \\
Frozen entrees & 2 & 91 & 204--218 & −8.4 & 9 & 6.6 & 1.0 to 12.1 & 9.7 \\
Frozen entrees & 6 & 40 & 205--218 & −7.6 & 2 & −3.2 & −6.8 to 0.4 & −2.1 \textsuperscript{*} \\
Frozen entrees & 12 & 114 & 205--218 & −8.7 & 3 & 9.2 & −5.7 to 24.0 & 7.5 \\
Frozen entrees & 2 & 119 & 205--218 & −8.0 & 9 & −0.1 & −5.6 to 5.4 & −1.1 \\
Frozen entrees & 1 & 2 & 220--237 & 9.0 & 4 & −6.2 & −11.0 to −1.5 & −12.8 \\
Frozen entrees & 5 & 8 & 220--237 & 9.8 & 2 & −13.0 & −95.9 to 69.9 & −7.7 \textsuperscript{*} \\
Frozen entrees & 2 & 9 & 220--237 & −9.6 & 9 & 13.3 & 9.4 to 17.2 & 7.8 \\
Frozen entrees & 7 & 33 & 220--237 & 5.9 & 2 & 2.3 & 1.0 to 3.6 & −1.1 \textsuperscript{*} \\
Frozen entrees & 6 & 40 & 220--237 & −10.9 & 1 & 21.0 & 21.0 to 21.0 & 15.6 \textsuperscript{*} \\
Frozen entrees & 2 & 49 & 220--234 & 8.9 & 9 & 4.7 & 0.8 to 8.6 & 5.0 \\
Frozen entrees & 1 & 68 & 220--237 & 7.6 & 4 & −3.1 & −7.8 to 1.7 & −12.2 \\
Frozen entrees & 1 & 71 & 220--237 & 10.0 & 4 & 7.4 & 2.7 to 12.2 & −5.4 \\
Frozen entrees & 7 & 75 & 220--234 & −10.2 & 2 & −0.3 & −1.6 to 1.0 & −4.7 \textsuperscript{*} \\
Frozen entrees & 2 & 76 & 220--234 & 8.6 & 9 & 4.7 & 0.8 to 8.6 & 4.8 \\
Frozen entrees & 1 & 95 & 220--234 & 8.1 & 4 & −1.7 & −6.4 to 3.1 & −6.4 \\
Frozen entrees & 2 & 110 & 220--234 & 8.4 & 9 & −3.7 & −7.6 to 0.2 & −1.6 \\
Frozen entrees & 2 & 119 & 220--237 & −6.6 & 9 & 2.0 & −1.9 to 5.9 & −2.8 \\
Frozen entrees & 1 & 130 & 220--233 & −7.3 & 4 & −3.3 & −8.0 to 1.5 & 2.1 \\
Frozen entrees & 1 & 71 & 270--290 & −7.0 & 14 & 9.4 & 6.0 to 12.8 & 5.0 \\
Soft drinks & 12 & 105 & 1--20 & −7.4 & 7 & 18.0 & 18.0 to 18.0 & 8.8 \textsuperscript{*} \\
Snack crackers & 2 & 50 & 196--216 & 9.6 & 10 & 9.4 & 3.2 to 15.6 & −13.1 \\
Snack crackers & 2 & 48 & 197--218 & 8.5 & 9 & −12.8 & −18.5 to −7.1 & −13.2 \\
Snack crackers & 2 & 76 & 197--218 & 6.9 & 9 & −11.9 & −18.1 to −5.8 & −6.8 \\
Snack crackers & 2 & 49 & 198--218 & 9.6 & 9 & −14.3 & −19.2 to −9.3 & −6.6 \\
Snack crackers & 2 & 89 & 198--218 & 8.5 & 9 & −7.7 & −12.6 to −2.8 & −1.1 \\
Snack crackers & 2 & 47 & 200--218 & 6.5 & 9 & 7.2 & 2.7 to 11.6 & 9.6 \\
Snack crackers & 2 & 28 & 203--218 & −6.5 & 9 & 24.0 & 20.2 to 27.8 & 22.4 \\
Snack crackers & 5 & 94 & 203--218 & −7.1 & 2 & −1.2 & −8.0 to 5.7 & −6.2 \textsuperscript{*} \\
Snack crackers & 1 & 2 & 204--218 & 8.7 & 4 & 5.5 & −7.4 to 18.3 & 2.9 \\
Snack crackers & 7 & 75 & 204--218 & −7.4 & 2 & −9.9 & −10.9 to −8.8 & −12.6 \textsuperscript{*} \\
Snack crackers & 2 & 88 & 204--218 & −8.6 & 9 & −2.9 & −7.2 to 1.3 & −0.6 \\
Snack crackers & 12 & 114 & 204--218 & −6.7 & 2 & 9.2 & −2.7 to 21.0 & 6.0 \textsuperscript{*} \\
Snack crackers & 2 & 119 & 204--218 & −6.5 & 9 & 3.5 & −0.8 to 7.7 & 4.1 \\
Snack crackers & 2 & 113 & 206--218 & −6.3 & 9 & 0.7 & −3.9 to 5.3 & 0.1 \\
Snack crackers & 6 & 40 & 207--218 & −8.7 & 2 & 9.5 & −18.8 to 37.8 & 14.6 \textsuperscript{*} \\
Snack crackers & 12 & 129 & 207--218 & −7.4 & 2 & 2.1 & −4.9 to 9.0 & −0.9 \textsuperscript{*} \\
\bottomrule
\end{tabular}

\end{table}

\section{The plan and its amendments}
\label{app:plan}

The plan (\texttt{docs/PAPER2\_PLAN.md}) was committed before any audit result
on 13 October 2026. It fixed the eight checks, the C7 detector, estimator and
per-episode score, and the rule that no audit result changes a model setting.
Eight dated entries followed. Entries 1 to 5 and 8 were made before the results
they affect were read; entries 6 and 7 were made after the earlier audit's
results, and everything they add is reported as added after the results.
\begin{enumerate}\setlength{\itemsep}{1pt}
\item 13 October, before any model lift: the episode detector measures each
store against its zone peers rather than the zone's most common price.
\item 13 October, after the episode list, before any model lift: at least three
controls to score an episode; models scored on an episode's first eight weeks,
against an 8-week difference in differences.
\item 13 October, before any audit result: the known-answer study, the C7
robustness analyses (event study, placebo, detector settings, pooled slope by
category), the pass thresholds for C1 to C6 and the main test.
\item 13 October, after the C7 robustness results, before any model lift for
C7: the price path and the pooled-slope intervals, labelled exploratory, and
the comparison of models with the measured response made on the pooled slope.
\item 13 October, correction, before any model lift for C7: an episode is
scored only if its 26 comparison weeks exist, leaving \zScored{} scored.
\item 17 October, after the audit results: the diagnosis of C1 and C4 on
fourteen saved runs; the main test also over versions; pass counts without C5,
and without C1 and C5, as exploratory.
\item 18 October, after the audit results: C1 under the model's own plan for
nothing planned; the comparison of the two plans for every configuration and
window; the benchmark described as observational; the main test rerun holding
the data and window fixed.
\item 13 November, before any audit of the tuned models: the audit moves to
each data set's final tuned world model; C1 scored under both plans; C3
between a model's copies; C6 against the tuned LightGBM; C7 and C8 with the
final settings; and the test of 18 October applied to the versions the
settings searches saved.
\item 16 November, after the results: C4 counted within the prices each item had in training; what a
wrong lift costs in the known-answer worlds (choosing half the planned weeks, ordering to the stated
lift); and the known-answer study rerun in eight worlds with one setting changed at a time.
\end{enumerate}

\section{Every known-answer model}
\label{app:known}

\begin{table}[H]
\centering
\caption{\textbf{Every model in the known-answer study, with its explanation
scored against the truth.} Lift against truth: rank agreement and ratio of
summed lifts on item-weeks whose true lift is not zero, overall and for the
price, promotion and display parts. C4: item-weeks lower after a deeper cut. C5: summed forecast over
summed sales on planned weeks. Passed: checks passed of those that apply (C1 to
C6). RMSSE: item by store, weeks 253 to 260.}
\label{tab:known}
\footnotesize
\begin{tabular}{@{}lrrrrrrrrr@{}}
\toprule
 & \multicolumn{5}{c}{Lift against truth} & & & & \\
\cmidrule(lr){2-6}
Model & rank & sum & price & promotion & display & C4 & C5 & Passed & RMSSE \\
\midrule
\multicolumn{10}{@{}l}{\emph{Plans at random}} \\
World model, seed 1 & 0.99 & 1.14 & 1.13 & 1.08 & 1.00 & 0.0\% & 1.08 & 6/6 & 0.401 \\
World model, seed 2 & 0.99 & 1.10 & 1.09 & 1.06 & 1.02 & 0.0\% & 1.06 & 6/6 & 0.401 \\
World model, seed 3 & 0.99 & 1.11 & 1.11 & 1.04 & 0.91 & 0.0\% & 1.05 & 5/6 & 0.398 \\
World model, blind to price & 0.59 & 0.73 & 0.01 & 2.00 & 0.92 & 31.5\% & 0.90 & 3/5 & 0.510 \\
World model, blind to levers & 0.89 & 1.10 & 1.35 & 0.14 & 0.05 & 0.0\% & 1.09 & 4/5 & 0.450 \\
LightGBM & 0.85 & 1.09 & 1.03 & 0.98 & 1.39 & 7.6\% & 0.99 & 1/2 & 0.389 \\
LightGBM, blind to price & 0.43 & 0.73 & −0.00 & 1.95 & 1.41 & 49.5\% & 0.88 & 1/2 & 0.515 \\
\multicolumn{10}{@{}l}{\emph{Plans in high-demand weeks}} \\
World model, seed 1 & 0.96 & 1.33 & 1.22 & 1.29 & 1.83 & 0.0\% & 1.06 & 5/6 & 0.345 \\
World model, seed 2 & 0.95 & 1.35 & 1.24 & 1.31 & 1.81 & 0.0\% & 1.07 & 6/6 & 0.349 \\
World model, seed 3 & 0.95 & 1.40 & 1.27 & 1.38 & 1.93 & 0.0\% & 1.09 & 5/6 & 0.344 \\
World model, blind to price & 0.51 & 1.09 & −0.00 & 2.37 & 2.05 & 27.2\% & 0.93 & 2/5 & 0.489 \\
World model, blind to levers & 0.82 & 1.23 & 1.63 & 0.06 & −0.03 & 0.0\% & 1.07 & 4/5 & 0.480 \\
LightGBM & 0.82 & 1.26 & 1.13 & 1.07 & 1.63 & 9.9\% & 1.01 & 1/2 & 0.322 \\
LightGBM, blind to price & 0.54 & 0.94 & −0.02 & 2.02 & 1.76 & 43.4\% & 0.90 & 1/2 & 0.497 \\
\bottomrule
\end{tabular}

\end{table}

\end{document}